\documentclass[runningheads]{llncs}
\usepackage{graphicx}
\usepackage{comment}
\usepackage{amsmath,amssymb} 
\usepackage{color}

\usepackage{epsfig}
\usepackage{graphicx}
\usepackage{amsmath}
\usepackage{amssymb}
\usepackage{booktabs}
\usepackage{comment}
\usepackage{subfigure}
\usepackage[normalem]{ulem}
\usepackage{multirow}

\usepackage{listings}
\usepackage{color}
\usepackage{floatrow}
\newfloatcommand{capbtabbox}{table}[][\FBwidth]

\definecolor{dkgreen}{rgb}{0,0.6,0}
\definecolor{gray}{rgb}{0.5,0.5,0.5}
\definecolor{mauve}{rgb}{0.58,0,0.82}

\usepackage{graphicx}
\usepackage{amsmath,amssymb} 

\usepackage[pagebackref=true,breaklinks=true,letterpaper=true,colorlinks,bookmarks=false]{hyperref}

\begin{document}

\title{Face Anti-Spoofing with Human Material Perception} 


\author{Zitong Yu\textsuperscript{1}, Xiaobai Li\textsuperscript{1}, Xuesong Niu \textsuperscript{2,3}, Jingang Shi\textsuperscript{4,1}, Guoying Zhao\textsuperscript{1\thanks{Corresponding author}}}
\institute{\small{\textsuperscript{1}Center for Machine Vision and Signal Analysis, University of Oulu, Finland  \\  \textsuperscript{2} Key Laboratory of Intelligent Information Processing of Chinese Academy of Sciences (CAS), Institute of
Computing Technology, CAS, China
\\
\textsuperscript{3} University of Chinese Academy of Sciences, China
\\
\textsuperscript{4} School of Software Engineering, Xi’an Jiaotong University, China
\\
\tt\scriptsize {\{zitong.yu, xiaobai.li, guoying.zhao\}@oulu.fi, \{xuesong.niu\}@vipl.ict.ac.cn}}}

\maketitle

\begin{abstract}

Face anti-spoofing (FAS) plays a vital role in securing the face recognition systems from presentation attacks. Most existing FAS methods capture various cues (e.g., texture, depth and reflection) to distinguish the live faces from the spoofing faces. All these cues are based on the discrepancy among physical materials (e.g., skin, glass, paper and silicone). In this paper we rephrase face anti-spoofing as a material recognition problem and combine it with classical human material perception~\cite{sharan2013recognizing}, intending to extract discriminative and robust features for FAS. To this end, we propose the Bilateral Convolutional Networks (BCN), which is able to capture intrinsic material-based patterns via aggregating multi-level bilateral macro- and micro- information. Furthermore, Multi-level Feature Refinement Module (MFRM) and multi-head supervision are utilized to learn more robust features. Comprehensive experiments are performed on six benchmark datasets, and the proposed method achieves superior performance on both intra- and cross-dataset testings. One highlight is that we achieve overall 11.3$\pm$9.5\% EER for cross-type testing in SiW-M dataset, which significantly outperforms previous results. We hope this work will facilitate future cooperation between FAS and material communities.

\end{abstract}

\vspace{-2.6em}
\section{Introduction}
\vspace{-0.8em}


In recent years, face recognition has been widely used in various interactive and payment scene due to its high accuracy and convenience. However, such biometric system is vulnerable to presentation attacks (PAs). Typical examples of physical presentation attacks include print, video replay, 3D masks and makeup. In order to detect such PAs and secure the face recognition system, face anti-spoofing (FAS) has attracted more attention from both academia and industry.

In the past decade, several hand-crafted feature based~\cite{boulkenafet2015face,Boulkenafet2017Face,Pereira2012LBP,Komulainen2014Context,Peixoto2011Face,Patel2016Secure} and deep learning based~\cite{qin2019learning,wang2020deep,yu2020searching,jourabloo2018face} methods have been proposed for presentation attack detection (PAD). On one hand, the classical hand-crafted descriptors leverage local relationship among the neighbours as the discriminative features, which is robust for describing the detailed invariant information (e.g., color texture, moir$\rm\acute{e}$ pattern and noise artifacts) between the live and spoofing faces. On the other hand, due to the stacked convolution operations with nonlinear activation, the convolutional neural networks (CNN) hold strong representation abilities to distinguish the bona fide from PA. However, most existing CNN and hand-crafted features are designed for universal image recognition tasks, which might not represent fine-grained spoofing patterns in FAS task.

\begin{figure}[t]
\vspace{-0.6em}
\centering
\includegraphics[width=10.5cm,height=4.6cm]{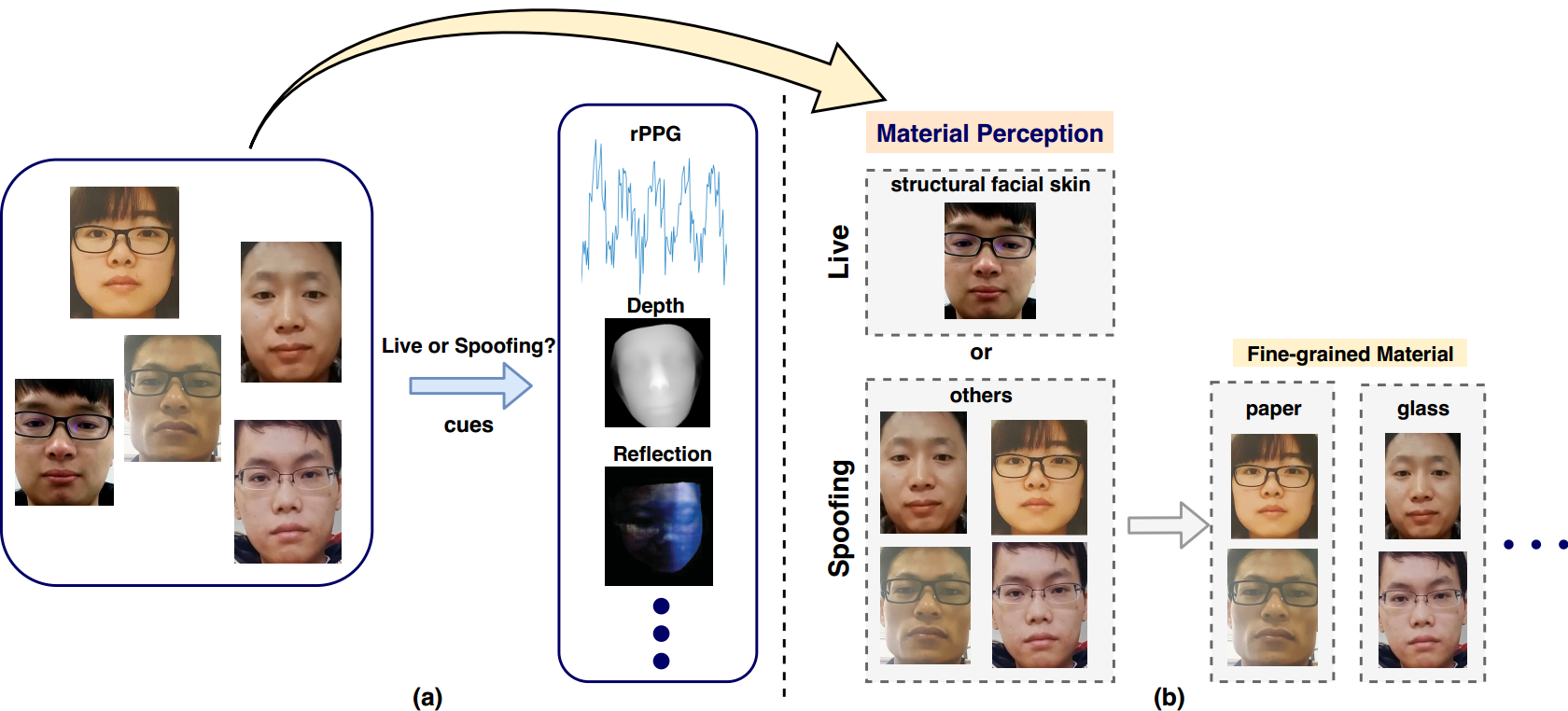}
  \caption{\small{
  (a) Face anti-spoofing can be regarded as a binary classification (live or spoofing) problem, which relies on the intrinsic cues such as rPPG, depth, reflection and so on. (b) Face anti-spoofing can be also treated as a material perception problem. }
  }
 
\label{fig:Figure1}
\vspace{-0.6em}
\end{figure}

According to the known intrinsic cues in face anti-spoofing task, many state-of-the-art methods introduced task-oriented priori knowledge for feature representation. As shown in Fig.~\ref{fig:Figure1}(a), there are three famous human-defined cues (i.e., rPPG, depth and reflection) for FAS task. Firstly, frequency distribution dissimilarity of rPPG signals ~\cite{lin2019face,li2016generalized,Liu2018Learning,liu2018remote} recovered from live skin surface and spoofing face can be utilized as there are no or weaker blood volume changes in spoofing faces. Secondly, structural facial depth difference between live and spoofing faces ~\cite{Liu2018Learning,Atoum2018Face} can be adopted as significant cue as most spoofing faces are broadcasted in plane presentation attack instruments
(PAIs). Thirdly, reflectance difference ~\cite{tan2010face,li20193d1} is also one kind of reliable cues as human facial skin and spoofing surfaces react differently to changes in illumination. Despite the human-defined cues are helpful to enhance the modeling capability respectively, it is still difficult to describe intrinsic and robust features for FAS task.

\textbf{An interesting and essential question for FAS task is how human beings differentiate live or spoofing faces, and what can be learned by machine intelligent systems?} In real-world cases, spoofing faces are always broadcasted by physical spoofing carriers (e.g., paper, glass screen and resin mask), which have obvious material properties difference with human facial skin. Such difference can be explicitly described as human-defined cues (e.g., rPPG, depth and reflection) or implicitly learned according to the material property uniqueness of structural live facial skin. Therefore, as illustrated in Fig.~\ref{fig:Figure1}(b), we assume that discrepancy of the structural materials between human facial skin and physical spoofing carriers are the essence of distinguishing live faces from spoofing ones.

Motivated by the discussions above, we rephrase face anti-spoofing task as structural material recognition problem and our goal is to learn intrinsic and robust features for distinguishing structural facial skin material from the others (i.e., materials for physical spoofing carriers). According to the study inspired by classical human material perception~\cite{sharan2013recognizing}, bilateral filtering plays a vital role in representing macro- and micro- cues for various materials. In this paper, we integrate traditional bilateral filtering operator into the state-of-the-art FAS deep learning framework, intending to help networks to learn more intrinsic material-based patterns. Our contributions include:

\begin{itemize}
\setlength\itemsep{-0.1em}
\vspace{-0.5em}
    \item We design novel Bilateral Convolutional Networks (BCN), which is able to capture intrinsic material-based patterns via aggregating multi-level bilateral macro- and micro- information.

    \item  We propose to use Multi-level Feature Refinement Module (MFRM) and material based multi-head supervision to further boost the performance of BCN. The former one refines the multi-scale features via reassembling weights of local neighborhood while the latter forces the network to learn robust shared features for multi-head auxiliary tasks.

    \item Our proposed method achieves outstanding  performance on six benchmark datasets with both intra- and cross-dataset testing protocols. We also conduct fine-grained material recognition experiments on SiW-M dataset to validate the effectiveness of our proposed method.

\end{itemize}

\vspace{-1.8em}
\section{Related Work}
\vspace{-0.5em}


\subsection{Face Anti-Spoofing}     
\vspace{-0.5em}
Traditional face anti-spoofing methods usually extract hand-crafted features from the facial images to capture the spoofing patterns. Several classical local descriptors such as LBP~\cite{boulkenafet2015face,Pereira2012LBP}, 
SIFT~\cite{Patel2016Secure}, SURF~\cite{Boulkenafet2017Face_SURF}, HOG~\cite{Komulainen2014Context} and DoG~\cite{Peixoto2011Face} are utilized to extract frame level features while video level methods usually capture dynamic cues like dynamic texture~\cite{komulainen2012face}, micro--motion~\cite{siddiqui2016face} and eye blinking~\cite{Pan2007Eyeblink}. More recently, a few deep learning based methods are proposed for FAS task. Some frame-level CNN methods \cite{yu2020auto,Li2017An,Patel2016Cross,george2019deep,jourabloo2018face,yu2020multi} are supervised by binary scalars or pixel-wise binary maps. In contrast, auxiliary depth~\cite{yu2020searching,Atoum2018Face,Liu2018Learning} and reflection~\cite{kim2019basn} supervisions are introduced to learn detailed cues effectively. In order to learn generalized features for unseen attacks and environment, few-shot learning~\cite{qin2019learning}, zero-shot learning~\cite{qin2019learning,liu2019deep} and domain generalization~\cite{jia2020single,wang2020cross,shao2019multi} are introduced for FAS task. Meanwhile, several video-level CNN methods are presented to exploit the dynamic spatio-temporal~\cite{wang2020deep,wang2018exploiting,yang2019face,lin2018live} or rPPG~\cite{li2016generalized,Liu2018Learning,lin2019face,yu2019remote,yu2020autohr,yu2019remote2} features for PAD. Despite introducing task-oriented priori knowledge (e.g., auxiliary depth, reflection and rPPG), deep learning based methods are still difficult to extract rich intrinsic features among live faces and various kinds of PAs.


\vspace{-1.0em}
\subsection{Human and Machine Material Perception} 
\vspace{-0.3em}
Our world consists of not only objects and scenes
but also of materials of various kinds. The perception of materials by humans usually focuses on optical and mechanical properties. Maloney and Brainard \cite{maloney2010color} demonstrates the research concerns about perception of material surface properties other than color and lightness, such as gloss or roughness. Fleming \cite{fleming2014visual} proposes statistical appearance models to describe visual perception of materials. Nishida \cite{nishida2019image} presents that material perception is visual estimation of optical modulation of image statistics. Inspired by human material perception, several machine intelligent methods are designed for material classification. Techniques derived from the domain of texture analysis can be adopted for material recognition by machines \cite{adelson2001seeing}. Varma and Zisserman \cite{varma2008statistical} utilizes joint distribution of intensity values over image patch exemplars for material classification under unknown viewpoint and illumination. Sharan et al. \cite{sharan2013recognizing} uses bilateral based low and mid-level image features for material recognition.  Aiming to keep the details of features, deep dilated convolutional network is used for material perception \cite{jiang2018deep}. 

In terms of vision applications, concepts of human material perception have been developed into image quality assessment \cite{ling2018role} and video quality assessment \cite{deng2016video}. For face anti-spoofing task, few works \cite{tan2010face,li20193d1,li20203d} consider discrepant surface reflectance properties of live or spoofing faces. However, only considering surface reflectance properties is not always reliable for material perception \cite{sharan2013recognizing}. In order to learn more generalized material-based features for FAS, we combine the state-of-the-art FAS methods with classical human material perception\cite{sharan2013recognizing}.


\vspace{-1.5em}
\section{Methodology}
\label{sec:method}
\vspace{-0.5em}

\begin{figure}[t]
\centering
\includegraphics[width=8.6cm,height=2.6cm]{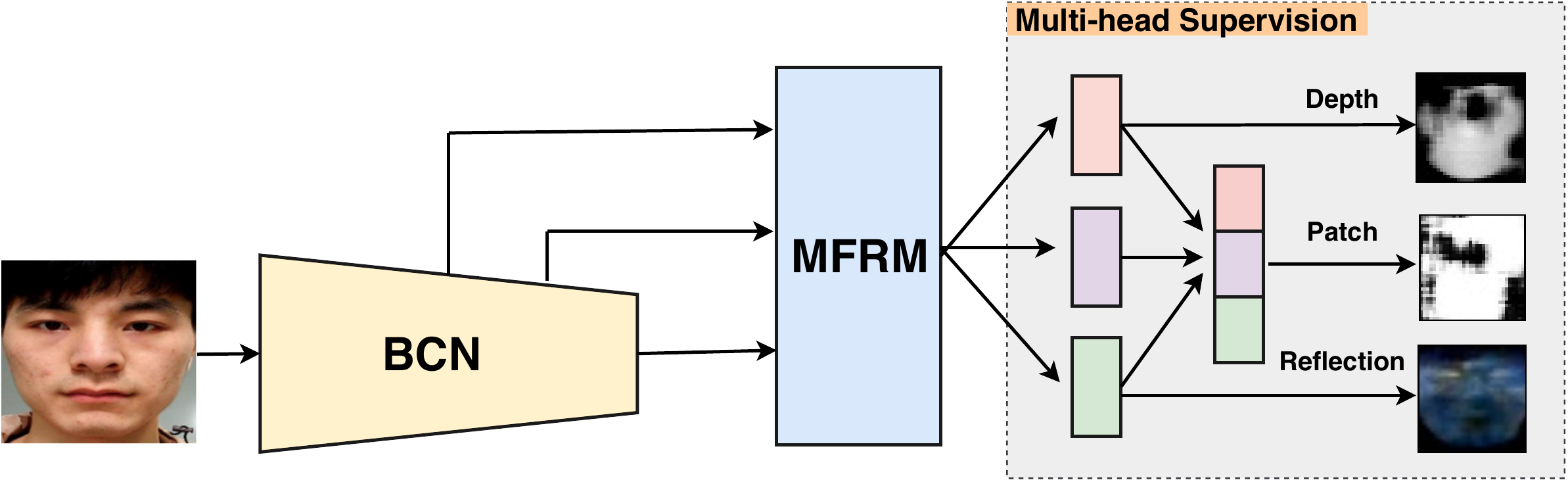}
  \caption{\small{
  The overall framework consists of Bilateral Convolutional Networks (BCN), Multi-level Feature Refinement Module (MFRM) and multi-head supervision. 
  }
  }
\label{fig:framework}
\vspace{-0.5em}
\end{figure}

In this section, we first introduce the Bilateral  Convolutional Networks (BCN) in Section~\ref{sec:BCN}, then present  Multi-level Feature Refinement Module (MFRM) in Section~\ref{sec:MFRM}, and at last introduce the material based multi-head supervision for face anti-spoofing in Section~\ref{sec:multihead}. The overall framework is shown in Fig.~\ref{fig:framework}.

\vspace{-0.8em}
\subsection{Bilateral Convolutional Networks}
\label{sec:BCN}
\vspace{-0.3em}

Inspired by classical material perception \cite{sharan2013recognizing} that utilizes bilateral filtering \cite{tomasi1998bilateral} for exacting subsequent macro- and micro- features, we try to adopt bilateral filtering technique for FAS task. The main issue is that in \cite{sharan2013recognizing}, several hand-crafted features are designed after bilateral filtering, which limits the feature representation capacity. In this subsection, we propose two solutions to integrate bilateral filtering with the state-of-the-art deep networks for FAS task.

\textbf{Bilateral Filtering.}\quad   
The first solution is straightforward: The bilateral filtered frames are taken as network inputs instead of the original RGB frames. The bilateral filter is utilized to smooth the original frame while preserving its main edges. Each pixel is a weighted mean of its neighbors where the weights decrease with the distance in space and with the intensity difference. With Gaussian function $g_{\sigma}(x))=exp(-x^{2}/\sigma ^{2})$, the bilateral filter of image $I$ at pixel $p$ is defined by:

\vspace{-2.2em}
\begin{equation} \small
\setlength{\belowdisplayskip}{-0.05em}
\begin{split}
&Bi\_Base(I)_{p}=\frac{1}{k}\sum_{q\in I}g_{\sigma_{s}}(\left \| p-q \right \|)g_{\sigma_{r}}(\left | I_{p}-I_{q} \right |)I_{q},
\\
&with: \qquad k=\sum_{q\in I}g_{\sigma_{s}}(\left \| p-q \right \|)g_{\sigma_{r}}(\left | I_{p}-I_{q} \right |),
\end{split}
\label{eq:bilateral}
\vspace{-2.5em}
\end{equation}
where $\sigma_{s}$ and $\sigma_{r}$ control the influence of spatial neighborhood distance and intensity difference respectively, and $k$ normalizes the weights. Give the input image $I$, bilateral filter is able to create a two-scale decomposition \cite{durand2002fast} where the output of the filter produces a large-scale base image $Bi\_Base(I)$ and the residual detail image $Bi\_Residual(I)$ can be obtained by $Bi\_Residual(I) = I-Bi\_Base(I)$. We use the fast approximation version\footnote{http://people.csail.mit.edu/jiawen/software/bilateralFilter-1.0.m} of the bilateral filter \cite{paris2006fast} with default parameters for implementation. 

\begin{figure}[t]
\centering
\includegraphics[width=12.5cm,height=2.5cm]{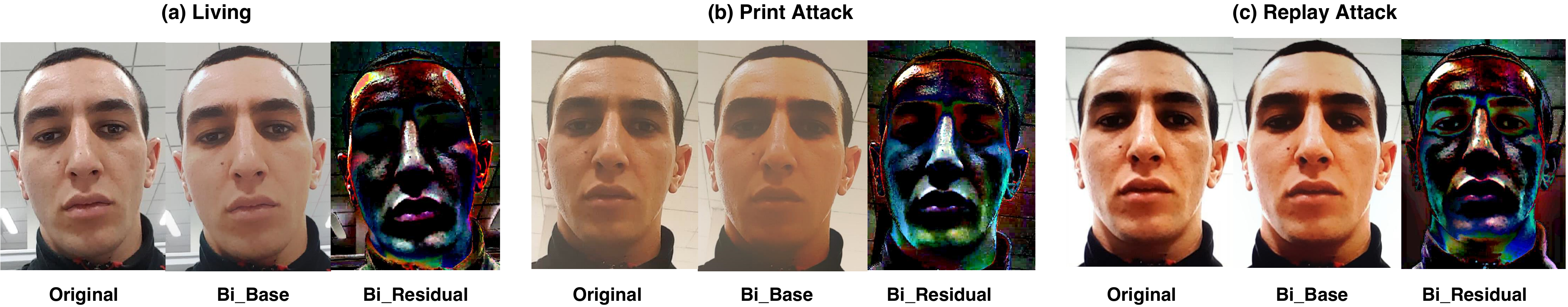}
  \caption{\small{
  Samples visualization for (a) live faces, (b) print attack, and (c) replay attack. `Bi\_Base' denotes frames after bilateral filtering while `Bi\_Residual' denotes residual result between original and bilateral filtered frames respectively. The intensity values of bilateral residual images are enlarged by four times for better visual effects.  
  }
  }
\label{fig:bilateral}
\vspace{-0.1em}
\end{figure}

Typical samples before and after bilateral filtering are visualized in Figure~\ref{fig:bilateral}. There are obvious differences in bilateral base and residual images between live and spoofing faces despite their similarities in the original RGB images. As shown in Fig.~\ref{fig:bilateral}(b) `Bi\_Base', the print attack face made of paper material is rougher and less glossy. Moreover, it can be seen from Fig.~\ref{fig:bilateral}(b)(c) `Bi\_Residual' that the high-frequency activation in eyes and eyebrow region is stronger, which might be caused by discrepant surface reflectance properties among materials (e.g., skin, paper and glass). These visual evidences are consistent with classical human material perception \cite{sharan2013recognizing} that macro- cues from bilateral base and micro- cues from bilateral residual are helpful for material perception.

In this paper, Auxiliary(Depth) \cite{Liu2018Learning} is chosen as our baseline deep model. The bilateral filtered (i.e., bilateral base and residual) images can forward the baseline model directly and predict the corresponding results. The ablation study of different kinds of inputs will be discussed in Section~\ref{sec:Ablation}.

\textbf{Deep Bilateral Networks.}\quad  The drawbacks of the above-mentioned solution are mainly of two folds: 1) directly replacing original inputs with bilateral images might lead to information loss, which limits the feature representation capability for neural networks, and 2) it is an inefficient way to learn multi-level bilateral features as the bilateral filter is only adopted in the input space. Aiming to overcome these drawbacks, we propose a novel method called Bilateral Convolutional Networks (BCN) to integrate traditional bilateral filtering with deep networks properly.

In order to filter the deep features instead of original images, the deep bilateral operator (DBO) is introduced. Mimicking the process of gray-scale or color image filtering, given the deep feature maps $\mathcal{F}\in \mathbb{R}^{H\times W\times C}$ with height $H$, width $W$ and $C$ channels, channel-wise deep bilateral filtering is operated. Considering the small spatial distance for the widely used convolution with $3\times3$ kernel, the distance decay term in Eqn.~(\ref{eq:bilateral}) can be removed (see \textsl{Appendix A} for corresponding ablation study), which is more efficient and lightweight when operating in deep hidden space. Hence deep bilateral operator for each channel of $\mathcal{F}$ can be formulated as    
\vspace{-0.5em}
\begin{equation} 
\setlength{\belowdisplayskip}{-0.1em}
\begin{split}
& DBO (\mathcal{F})_{p}=\frac{1}{k}\sum_{q\in \mathcal{F}}g_{\sigma_{r}}(\left | \mathcal{F}_{p}-\mathcal{F}_{q} \right |)\mathcal{F}_{q},
\\
&with: \qquad k=\sum_{q\in \mathcal{F}}g_{\sigma_{r}}(\left | \mathcal{F}_{p}-\mathcal{F}_{q} \right |).
\end{split}
\label{eq:DBO}
\vspace{-0.2em}
\end{equation}

\begin{figure}[t]
\centering
\includegraphics[scale=0.48]{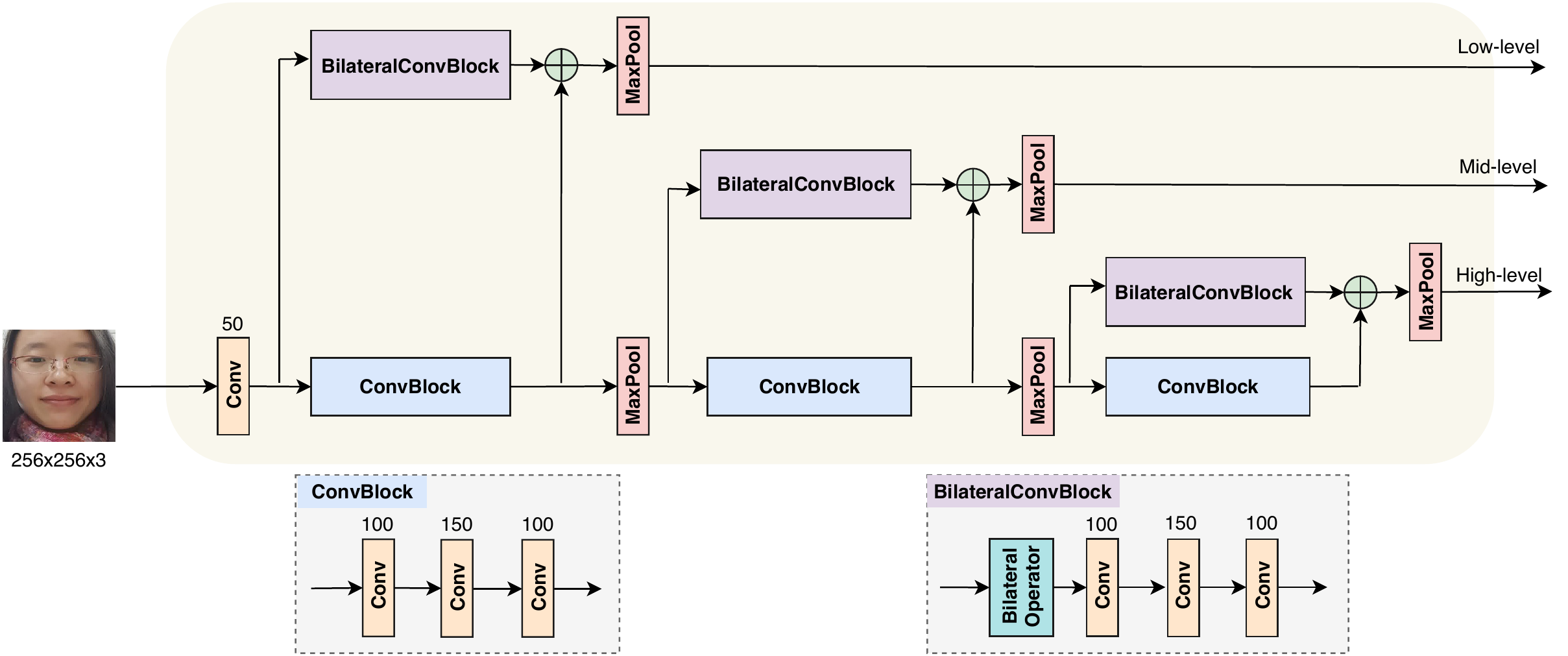}
  \caption{\small{
  The proposed BCN architecture. The number of filters are shown on top of each convolutional layer, the size of all filters is 3$\times$3 with stride 1 for convolutional and 2 for pooling layers. Each output from `ConvBlock' and `BilateralConvBlock' in the same level will be operated with element-wise addition. 
  }
}
\label{fig:BCN}
\vspace{-0.5em}
\end{figure}

Now performing DBO for features in different levels, it is easy to obtain multi-level bilateral base features. Nevertheless, how to get multi-level bilateral residual features is still unknown. As our goal is to represent aggregated bilateral base and residual features $\mathcal{F}_{Bi}$, inspired by residual learning in ResNet~\cite{he2016deep}, bilateral residual features $\mathcal{F}_{Residual}$ can be learned dynamically via shortcut connecting with bilateral base features $\mathcal{F}_{Base}$, i.e., $\mathcal{F}_{Residual}=\mathcal{F}_{Bi}-\mathcal{F}_{Base}$. 
 The architecture of the proposed Bilateral Convolutional Networks  is illustrated in Figure~\ref{fig:BCN}. As `BilateralConvBlock' and `ConvBlock' have same convolutional structure but unshared parameters, it is possible to learn $\mathcal{F}_{Base}$ and $\mathcal{F}_{Residual}$ from `BilateralConvBlock' and `ConvBlock' respectively. Compared with baseline model Auxiliary(Depth) \cite{Liu2018Learning} without `BilateralConvBlock', BCN is able to learn more intrinsic features via aggregating multi-level bilateral macro- and micro- information.

\subsection{Multi-level Feature Refinement Module}
\label{sec:MFRM}

In the baseline model Auxiliary(Depth) \cite{Liu2018Learning}, multi-level features are concatenated directly for subsequent head supervision. We argue that such coarse features are not optimal for fine-grained material-based FAS task. Hence Multi-level Feature Refinement Module (MFRM) is introduced after BCN (see Fig.~\ref{fig:framework} ), which aims to refine and fuse the coarse low-mid-high level features from BCN via context-aware feature reassembling.

\begin{figure}[t]
\centering
\includegraphics[scale=0.27]{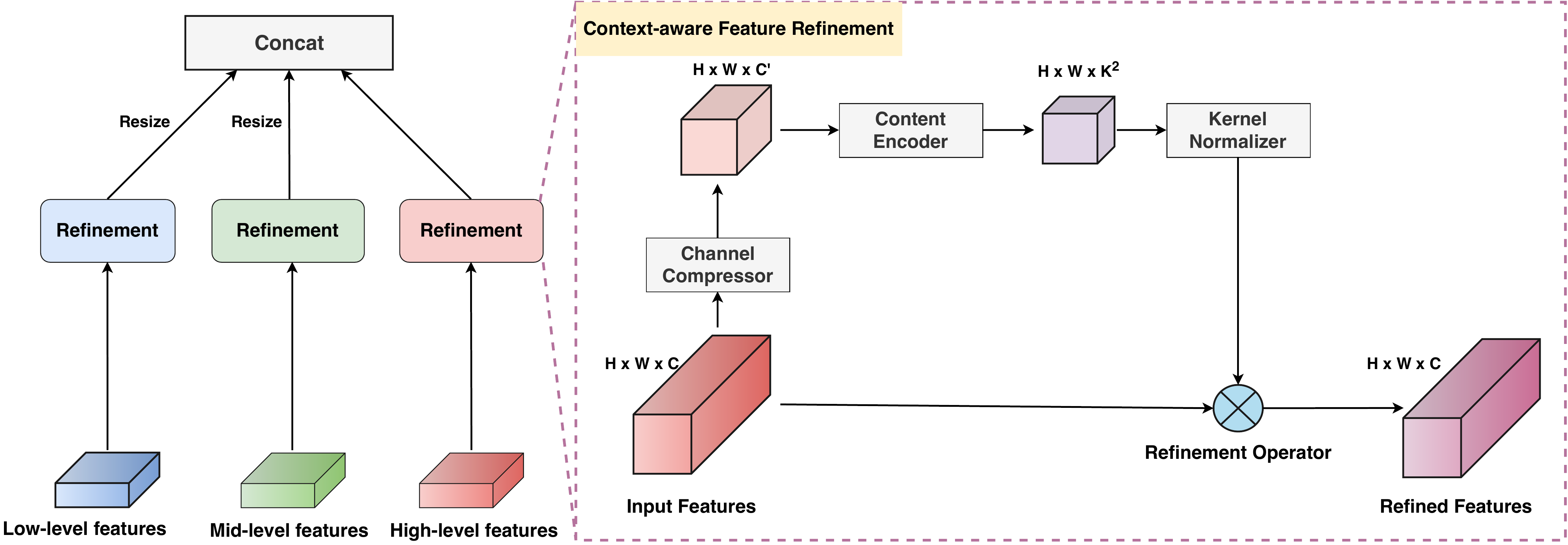}
  \caption{\small{
  Multi-level Feature Refinement Module. 
  }
  }
\label{fig:MFRM}
\vspace{-0.5em}
\end{figure}

As illustrated in Fig.~\ref{fig:MFRM}, features $\mathcal{F}$ from low-mid-high levels are refined via reassembling features with local context-aware weights. Unlike \cite{wang2019carafe} which aims for feature upsampling, here we focus on the general multi-level feature refinement. The refined features $\mathcal{F}'$ can be formulated as

\vspace{-0.5em}
\begin{equation} 
\mathcal{F}'_{level}=\mathcal{F}_{level}\, \otimes \, \mathcal{N}(\ \psi (\phi (\mathcal{F}_{level}))),\ level\in \left \{low, mid, high \right \},
\label{eq:refinement}
\end{equation}
where $\phi$, $\psi$, $\mathcal{N}$ and $\otimes$ represent channel compressor, content encoder, kernel normalizer and refinement operator, respectively. The channel compressor adopts a 1$\times$1 convolution layer to compress the input feature channel from $C$ to $C'$, making the refinement module more efficient. The content encoder utilizes a convolution layer of kernel size 5$\times$5 to generate refinement kernels based on the content of input features $\mathcal{F}$, and then each $K\times K$ refinement kernel is normalized with a softmax function spatially in kernel normalizer. Given the location $l=(i,j)$, channel $c$ and corresponding normalized refinement kernel $\mathcal{W}_{l}$, the output refined features $\mathcal{F}'$ are expressed as
\vspace{-0.6em}
\begin{equation} \small
\setlength{\belowdisplayskip}{-0.1em}
\mathcal{F}'_{(i,j,c)}=\sum_{n=-r}^{r}\sum_{m=-r}^{r}\mathcal{W}_{l(n,m)} \cdot \mathcal{F}_{(i+n,j+m,c)}, \ with \ r=\left \lfloor K/2 \right \rfloor.
\label{eq:refinement2}
\end{equation}
In essence, MFRM exploits the semantic and contextual information to reallocate the contributions of the local neighbors, which is possible to obtain more intrinsic features. For instance, our module is able to refine the discriminative cues (e.g. moir$\rm\acute{e}$ pattern) from salient regions according to their local context. We also compare the refinement method with other classical feature attention based methods such as spatial attention \cite{woo2018cbam}, channel attention \cite{hu2018squeeze} and non-local attention \cite{wang2018non} in Sec. \ref{sec:Ablation}.

\vspace{-1.0em}
\subsection{Material based Multi-head Supervision}
\label{sec:multihead}
\vspace{-0.5em}
As material categorization is complex and depends on various fine-grained cues (e.g., surface shape, reflectance and texture), it is impossible to learn robust and intrinsic material-based patterns via only one simple supervision (e.g., softmax binary loss and depth regression loss). For the sake of learning intrinsic material-based features, material based multi-head supervision is proposed. As shown in Fig.~\ref{fig:framework}, three-head supervision is utilized to guide the multi-level fused features from MFRM: 1) depth-head supervision, intending to force the model to learn structural surface shape information; 2) reflection-head supervision, helping networks to learn surface reflectance property; and 3) patch-head supervision, guiding to learn fine-grained surface texture cues. The detailed network structure can be found in \textit{Appendix B}.

\textbf{Loss Function.}\quad
Appropriate loss functions should be designed to supervise the network training. Given an input face image $I$, the network predicts the depth map $D_{pre}$, reflection map $R_{pre}$ and patch map $P_{pre}$. Then the loss functions can be formulated as:  

\vspace{-1.0em}
\begin{equation} \footnotesize
\setlength{\belowdisplayskip}{-0.1em}
\mathcal{L}_{depth}=\frac{1}{H\times W}\sum_{i\in{H},j\in{W}}\left \| D_{pre(i,j)} - D_{gt(i,j)}\right \|_{2}^{2},
\vspace{-0.2em}
\end{equation}
\begin{equation} \footnotesize
\mathcal{L}_{reflection}=\frac{1}{H\times W\times C}\sum_{i\in{H},j\in{W},c\in{C}}\left \| R_{pre(i,j,c)} - R_{gt(i,j,c)}\right \|_{2}^{2},
\vspace{-0.2em}
\end{equation}
\begin{equation} \footnotesize
\mathcal{L}_{patch}=\frac{1}{H\times W}\sum_{i\in{H},j\in{W}}-(P_{gt(i,j)}log(P_{pre(i,j)})+(1-P_{gt(i,j)})log(1-P_{pre(i,j)})),
\vspace{-0.2em}
\end{equation}
where $D_{gt}$, $R_{gt}$ and $P_{gt}$ denote ground truth depth map, reflection map and patch map respectively. Finally, the overall loss function is $\mathcal{L}_{overall} = \mathcal{L}_{depth}+\mathcal{L}_{reflection}+\mathcal{L}_{patch}$.

\textbf{Ground Truth Generation.}\quad
 Dense face alignment PRNet ~\cite{Feng2018Joint} is adopted to estimate facial 3D shapes and generate the facial depth maps with size $32\times32$. The reflection maps are estimated by the state-of-the-art reflection estimation network ~\cite{zhang2018single} and then face regions are cropped from
reflection maps to avoid being overfitted to backgrounds. The generated reflection maps have size with $32\times32\times3$ (3 channels for RGB). More details and samples can be found in ~\cite{wang2018exploiting,kim2019basn}. To distinguish live faces from spoofing faces, at the training stage, we normalize live depth maps and spoofing reflection maps in a range of $[0, 1]$, while setting spoofing depth maps and live reflection maps to 0, which is similar to ~\cite{Liu2018Learning,kim2019basn}. The patch maps are generated simply by downsampling original images and filling each patch position with corresponding
binary label (i.e., live 1 and spoofing 0). The generated binary patch maps keep size with $32\times32$.

\vspace{-0.4em}
\section{Experiments}
\vspace{-0.1em}
\label{sec:experiemnts}
In this section, comprehensive experiments are performed to evaluate our method. We will sequentially
describe the employed datasets \& metrics (Sec. \ref{sec:dataset}), implementation details (Sec. \ref{sec:Details}), results (Sec. \ref{sec:Ablation} - \ref{sec:Inter}) and analysis (Sec. \ref{sec:Analysis}).

\vspace{-0.1em}
\subsection{Datasets and Metrics}
\vspace{-0.1em}

\label{sec:dataset}
Six databases including OULU-NPU ~\cite{Boulkenafet2017OULU}, SiW ~\cite{Liu2018Learning}, CASIA-MFSD ~\cite{Zhang2012A}, Replay-Attack ~\cite{ReplayAttack}, MSU-MFSD ~\cite{wen2015face} and SiW-M ~\cite{liu2019deep} are used in our experiments. OULU-NPU and SiW are large-scale high-resolution databases, containing four and three protocols to validate the generalization (e.g., unseen illumination and attack medium) of models respectively, which are utilized for intra testing. CASIA-MFSD, Replay-Attack and MSU-MFSD are databases which contain low-resolution videos, and are used for cross testing. SiW-M is designed for fine-grained material recognition and cross-type testing for unseen attacks as there are rich attacks types (totally 13 types) inside.  

\textbf{Performance Metrics.}\quad
In OULU-NPU and SiW dataset, we follow the original protocols and metrics, i.e., Attack Presentation Classification Error Rate (APCER), Bona Fide Presentation Classification Error Rate (BPCER), and ACER ~\cite{ACER} for a fair comparison. Half Total Error Rate (HTER) is adopted in the cross testing between CASIA-MFSD and Replay-Attack. Area Under Curve (AUC) is utilized for intra-database cross-type test on CASIA-MFSD, Replay-Attack and MSU-MFSD. For the cross-type test on SiW-M, APCER, BPCER, ACER and Equal Error Rate (EER) are employed.

\vspace{-1.2em}
\subsection{Implementation Details}
\vspace{-0.2em}
\label{sec:Details}

Our proposed method is implemented with Pytorch. The default settings $\sigma_{r}=1.0$ and $C'=20, K=5$ are adopted for BCN and MFRM, respectively. In the training stage, models are trained with Adam optimizer and the initial learning rate (lr) and weight decay (wd) are 1e-4 and 5e-5, respectively. We train models with maximum 1300 epochs while lr halves every 500 epochs. The batch size is 7 on a Nvidia P100 GPU. In the testing stage, we calculate the mean value of the predicted depth map $\mathcal{D}_{test}$, reflection map $\mathcal{R}_{test}$ and patch map $\mathcal{P}_{test}$ as the final score $\mathcal{S}_{test}$: 
\vspace{-0.2em}
\begin{equation} 
\mathcal{S}_{test}=mean(\mathcal{D}_{test})+mean(1-\mathcal{R}_{test})+mean(\mathcal{P}_{test}).
\end{equation}
\vspace{-1.2em}

\begin{figure}[t]
\includegraphics[width=12.4cm,height=3.4cm]{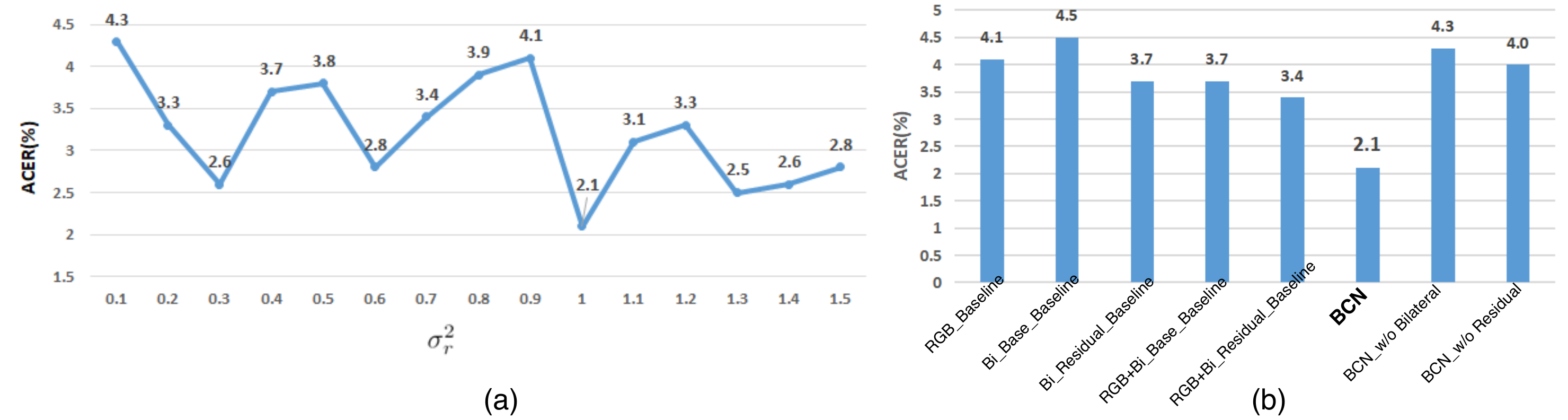}
  \caption{\small{
  (a) Impact of $\sigma_{r}$ in BCN. (b) Comparison among various kinds of inputs for baseline model Auxilary(Depth) ~\cite{Liu2018Learning} and BCN. Lower ACER, better performance.}
  }

\label{fig:Ablation1}
\vspace{-1.1em}
\end{figure}

\vspace{-1.8em}
\subsection{Ablation Study}
\vspace{-0.4em}
\label{sec:Ablation}
In this subsection, all ablation studies are conducted on Protocol-1 (different illumination condition and location between train and test sets) of OULU-NPU~\cite{Boulkenafet2017OULU} to explore the details of our proposed BCN, MFRM and multi-head supervision.

\begin{table}[t]
\begin{floatrow}

\capbtabbox{
\scalebox{0.81}{\begin{tabular}{|l|c |c|}
\hline
Model & ACER(\%)\\
\hline
D (Baseline) & 4.1 \\
D+R  & 2.8 \\
D+P  & 2.3 \\
D+R+P & 1.8 \\
D+R+P+MFRM  & 1.2 \\
D+R+P+MFRM+BCN  & 0.8 \\
\hline
\end{tabular}}
}{
 \caption{Results of network composition and supervision.}
 \label{tab:abnetwork}
}
\capbtabbox{
 
\scalebox{0.81}{\begin{tabular}{|l|c |c|}
\hline
Model & ACER(\%)\\
\hline
D+R+P & 1.8 \\
D+R+P+Spatial Attention \cite{woo2018cbam}  & 1.5 \\
D+R+P+Channel Attention \cite{hu2018squeeze} & 9.5 \\
D+R+P+Non-local Attention \cite{wang2018non} & 12.7 \\
D+R+P+Context-aware Reassembling  & 1.2 \\
\hline
\end{tabular}}
}{
 \caption{Ablation study of refinement methods in MFRM.}
 \label{tab:abMFRM}
}

\end{floatrow}
\end{table}

\textbf{Impact of $\sigma_{r}$ and Bilateral Operator in BCN.}\quad 
According to Eqn.~(\ref{eq:DBO}) and Gaussian function $g_{\sigma_{r}}(x))=exp(-x^{2}/\sigma_{r}^{2})$, $\sigma_{r}$ controls the strength of neighbor feature differences, i.e., the higher $\sigma_{r}$, the more contributions are given to the neighbor with large differences. As illustrated in Fig.~\ref{fig:Ablation1}(a), the best performance (ACER=2.1\%) is obtained when $\sigma_{r}^{2}=1.0$ in BCN. We use this setting for the following experiments. According to the quantitative index shown in Fig.~\ref{fig:Ablation1}(b), BCN (ACER=2.1\%) could decrease the ACER by half when campared with the results of `RGB\_Baseline' (ACER=4.1\%). In order to validate whether the improvement from BCN is due to the extra parameters from `BilateralConvBlock' in Fig.~\ref{fig:BCN}, we remove the bilateral operator in these blocks. However, as shown in Fig.~\ref{fig:Ablation1}(b) `BCN\_w/o Bilateral', it even works worse than the baseline (4.3\% versus 4.1\% ACER), indicating that the network easily overfits by introducing extra parameters without bilateral operators. We are also curious about the efficacy of bilateral residual term. After removing bilateral residual structure from BCN, the ACER index sharply changes from 2.1\% (see `BCN') to 4.0\% (see `BCN\_w/o Residual'). It implies that the micro- patterns from bilateral residual branch are also important for FAS task. 


\textbf{Influence of Various Input Types.}\quad 
As discussed in Sec. \ref{sec:BCN}, the first solution is to adopt bilateral filtered images as network inputs. The results in Fig.~\ref{fig:Ablation1}(b) show that `Bi\_Residual\_Baseline' with bilateral residual inputs performs better (0.4\% ACER lower) than that of original RGB baseline. While combing the multi-level features from both original RGB and bilateral filtered images (i.e., `RGB+Bi\_Base\_Baseline' and `RGB+Bi\_Residual\_Baseline' in Fig.~\ref{fig:Ablation1}(b)), the performance further boosts. 
In contrast, BCN with only original RGB input outperforms the first solution methods for a large margin, implying that deep bilateral base and residual features in BCN are more robust.

\textbf{Advantage of MFRM and Multi-head Supervision.}\quad 
Table~\ref{tab:abnetwork} shows the ablation study about the network composition and supervision. `D', `R', `P' are short for depth, reflection and patch heads, respectively. It is clear that multi-head supervision facilitates the network to learn more intrinsic features thus boost the performance. Furthermore, with both MFRM and multi-head supervision, our model is able to reduce ACER from baseline 4.1\% to 1.2\%. Ultimately, the full version of our method `D+R+P+MFRM+BCN' achieves excellent performance with 0.8\% ACER.

\textbf{Impact of Refinement Methods in MFRM.}\quad 
We investigate four feature refinement methods in MFRM and the results are shown in Table~\ref{tab:abMFRM}. It is surprised that only spatial attention \cite{woo2018cbam} (the second row) boosts the performance while SE block based channel attention \cite{hu2018squeeze} (the third row) and non-local block based self-attention\cite{wang2018non} (the fourth row) perform poorly when domain shifts (e.g., illumination changes). We adopt context-aware reassembling as defaulted setting in MFRM as it can obtain more generalized features and improve baseline `D+R+P' by 0.6\% ACER. In summary, we use `D+R+P+MFRM+BCN (with $\sigma_{r}^{2}=1.0$)' for all the following tests.

\begin{table}[t]
\centering
\caption{The results of intra testing on four protocols of OULU-NPU.}
\resizebox{0.55\textwidth}{!}{
\begin{tabular}{|c|c|c|c|c|}

\hline
Prot. & Method & APCER(\%)$\downarrow$ & BPCER(\%)$\downarrow$ & ACER(\%)$\downarrow$ \\
\hline
\multirow{6}{*}{1}
        &GRADIANT ~\cite{boulkenafet2017competition}&1.3 &12.5 & 6.9 \\
        &BASN ~\cite{kim2019basn}&1.5 &5.8 & 3.6 \\
        &STASN ~\cite{yang2019face} &1.2 &2.5 & 1.9 \\
        &Auxiliary ~\cite{Liu2018Learning} &1.6 &1.6 & 1.6 \\
        &FaceDs ~\cite{jourabloo2018face} &1.2 &1.7 & 1.5 \\
        &FAS-TD ~\cite{wang2018exploiting} &2.5 &0.0 & 1.3 \\
        &DeepPixBiS ~\cite{george2019deep}&0.8 &0.0 & \textbf{0.4} \\
        &\textbf{Ours} &0.0 &1.6 & \underline{0.8} \\
\hline
\multirow{6}{*}{2} 
       &DeepPixBiS ~\cite{george2019deep}&11.4 &0.6 & 6.0 \\
       &FaceDs ~\cite{jourabloo2018face}&4.2 &4.4 & 4.3 \\
       &Auxiliary ~\cite{Liu2018Learning}&2.7 &2.7 & 2.7 \\
       &BASN ~\cite{kim2019basn}&2.4 &3.1 & 2.7 \\
       &GRADIANT ~\cite{boulkenafet2017competition}&3.1 &1.9 & 2.5 \\
       &STASN ~\cite{yang2019face}&4.2 &0.3 & 2.2 \\
       &FAS-TD ~\cite{wang2018exploiting} &1.7 &2.0 & \underline{1.9} \\
        &\textbf{Ours} &2.6 &0.8 & \textbf{1.7} \\
\hline
\multirow{4}{*}{3} 
       &DeepPixBiS ~\cite{george2019deep}&11.7$\pm$19.6 &10.6$\pm$14.1 & 11.1$\pm$9.4 \\
       &FAS-TD ~\cite{wang2018exploiting}&5.9$\pm$1.9 &5.9$\pm$3.0 & 5.9$\pm$1.0 \\
       &GRADIANT ~\cite{boulkenafet2017competition}&2.6$\pm$3.9 &5.0$\pm$5.3 &3.8$\pm$2.4 \\
       &BASN ~\cite{kim2019basn}&1.8$\pm$1.1 &3.6$\pm$3.5 &2.7$\pm$1.6 \\
       &FaceDs ~\cite{jourabloo2018face}&4.0$\pm$1.8 &3.8$\pm$1.2 &3.6$\pm$1.6 \\
       &Auxiliary ~\cite{Liu2018Learning}&2.7$\pm$1.3 &3.1$\pm$1.7 &{2.9}$\pm$1.5 \\
       &STASN ~\cite{yang2019face}&4.7$\pm$3.9 &0.9$\pm$1.2  &\underline{2.8$\pm$1.6} \\
        &\textbf{Ours} &2.8$\pm$2.4 &2.3$\pm$2.8  & \textbf{2.5$\pm$1.1} \\
\hline
\multirow{4}{*}{4} 
        &DeepPixBiS ~\cite{george2019deep}&36.7$\pm$29.7 &13.3$\pm$14.1 & 25.0$\pm$12.7 \\
       &GRADIANT ~\cite{boulkenafet2017competition}&5.0$\pm$4.5 &15.0$\pm$7.1 &10.0$\pm$5.0 \\
       &Auxiliary ~\cite{Liu2018Learning}&9.3$\pm$5.6 &10.4$\pm$6.0 &9.5$\pm$6.0 \\
       &FAS-TD ~\cite{wang2018exploiting}&14.2$\pm$8.7 &4.2$\pm$3.8 & 9.2$\pm$3.4 \\
       &STASN ~\cite{yang2019face}&6.7$\pm$10.6 &8.3$\pm$8.4  &7.5$\pm$4.7 \\
       &FaceDs ~\cite{jourabloo2018face}&1.2$\pm$6.3 &6.1$\pm$5.1 &5.6$\pm$5.7 \\
       &BASN ~\cite{kim2019basn}&6.4$\pm$8.6 &3.2$\pm$5.3 &\textbf{4.8$\pm$6.4} \\
       &\textbf{Ours} &2.9$\pm$4.0 &7.5$\pm$6.9  &\underline{5.2$\pm$3.7} \\

\hline
\end{tabular}
}
\label{tab:OULU}
\vspace{-2.0em}
\end{table}

 \vspace{-1.4em}
\subsection{Intra Testing}
 \vspace{-0.5em}
 The intra testing is carried out on both the OULU-NPU
and the SiW datasets. We strictly follow the four protocols on OULU-NPU and three protocols on SiW for
the evaluation. All compared methods including STASN ~\cite{yang2019face} are trained without extra datasets for a fair comparison. 

\textbf{Results on OULU-NPU.} \quad  As shown in Table~\ref{tab:OULU}, our proposed method ranks first or second on all the four protocols (0.4\%, 1.7\%, 2.5\% and 5.2\% ACER, respectively), which indicates the proposed method performs well at the generalization of the external environment, attack mediums and input camera variation. Note that DeepPixBis ~\cite{george2019deep} utilizes patch map for supervision but performing poorly in unseen attack mediums and camera types while BASN ~\cite{george2019deep} exploits depth and reflection map as guidance but ineffective in unseen external environment. Our method works well for all 4 protocols as the extracted material-based features are intrinsic and generalized.

\textbf{Results on SiW.} \quad   We also compare our method with four state-of-the-art methods~\cite{Liu2018Learning,yang2019face,wang2018exploiting,kim2019basn} on SiW dataset. Our method performs the best for all three protocols (0.36\%, 0.11\%, 2.45\% ACER, respectively), revealing the excellent generalization capacity for 1) variations of face pose and expression, 2) variations of different spoof mediums, and 3) unknown presentation attack. More detailed results of each protocol are shown in \textsl{Appendix C}.

 \vspace{-0.4em}
\subsection{Inter Testing}
 \vspace{-0.3em}
\label{sec:Inter}

To further validate whether our model is able to learn intrinsic features, we conduct cross-type and cross-dataset testing to verify the generalization capacity to unknown presentation attacks and unseen environment, respectively.

\textbf{Cross-type Testing.} \quad
Here we use CASIA-MFSD~\cite{Zhang2012A}, Replay-Attack~\cite{ReplayAttack} and MSU-MFSD~\cite{wen2015face} to perform intra-dataset cross-type testing between replay and print attacks. Our proposed method achieves the best overall performance (96.77\% AUC) among state-of-the-art methods~\cite{arashloo2017anomaly,xiong2018unknown,liu2019deep,qin2019learning}, indicating the learned features generalized well among unknown attacks. More details can be found in \textsl{Appendix D}. Moreover, we also conduct cross-type testing on the latest SiW-M~\cite{liu2019deep} dataset. As illustrated in Table~\ref{tab:SiW-M}, the proposed method achieves the best average ACER (11.2\%) and EER (11.3\%) among 13 attacks, which indicates our method actually learns material-based intrinsic patterns from rich kinds of material hence generalized well in unseen material type.

\newcommand{\tabincell}[2]{\begin{tabular}{@{}#1@{}}#2\end{tabular}}
\begin{table}[t]
\centering
\vspace{-0.2em}
\caption{The evaluation and comparison of the cross-type testing on SiW-M~\cite{liu2019deep}.}
\vspace{-0.2em}
\scalebox{0.55}{\begin{tabular}{c|c|c|c|c|c|c|c|c|c|c|c|c|c|c|c}
\hline
\multirow{2}{*}{Method} &\multirow{2}{*}{Metrics(\%)} &\multirow{2}{*}{Replay} &\multirow{2}{*}{Print} &\multicolumn{5}{c|}{Mask Attacks} &\multicolumn{3}{c|}{Makeup Attacks}&\multicolumn{3}{c|}{Partial Attacks} &\multirow{2}{*}{Average} \\
\cline{5-15} &  &  &  & \tabincell{c}{Half} &\tabincell{c}{Silicone} &\tabincell{c}{Trans.} &\tabincell{c}{Paper}&\tabincell{c}{Manne.}&\tabincell{c}{Obfusc.}&\tabincell{c}{Imperson.}&\tabincell{c}{Cosmetic}&\tabincell{c}{Funny Eye} & \tabincell{c}{Paper Glasses} &\tabincell{c}{Partial Paper} & \\
\hline
\hline

\multirow{4}{*}{SVM+LBP~\cite{Boulkenafet2017OULU}} & APCER & 19.1 & 15.4 & 40.8 & 20.3 & 70.3 & 0.0 & 4.6 & 96.9 & 35.3 & 11.3 & 53.3 & 58.5 & 0.6 & 32.8$\pm$29.8 \\
\cline{3-16}  & BPCER & 22.1 & 21.5 & 21.9 & 21.4 & 20.7 & 23.1 & 22.9 & 21.7 & 12.5 & 22.2 & 18.4 & 20.0 & 22.9 & 21.0$\pm$2.9 \\
\cline{3-16}  & ACER & 20.6 & 18.4 & 31.3 & 21.4 & 45.5 & 11.6 & 13.8 & 59.3 & 23.9 & 16.7 & 35.9 & 39.2 & 11.7 & 26.9$\pm$14.5 \\
\cline{3-16}  & EER & 20.8 & 18.6 & 36.3  & 21.4 & 37.2 & 7.5 & 14.1 & 51.2 & 19.8 & 16.1 & 34.4 & 33.0 & 7.9 & 24.5$\pm$12.9 \\

\hline
\hline

\multirow{4}{*}{Auxiliary~\cite{Liu2018Learning}} & APCER & 23.7 & 7.3 & 27.7 & 18.2 & 97.8 & 8.3 & 16.2 & 100.0 & 18.0 & 16.3 & 91.8 & 72.2 & 0.4 & 38.3$\pm$37.4 \\
\cline{3-16}  & BPCER & 10.1 & 6.5 & 10.9 & 11.6 & 6.2 & 7.8 & 9.3 & 11.6 & 9.3 & 7.1 & 6.2 & 8.8 & 10.3 & 8.9$\pm$ 2.0 \\
\cline{3-16}  & ACER & 16.8 & 6.9 & 19.3 & 14.9 & 52.1 & 8.0 & 12.8 & 55.8 & 13.7 & \textbf{11.7} & 49.0 & 40.5 & 5.3 & 23.6$\pm$18.5 \\
\cline{3-16}  & EER & 14.0 & 4.3 & 11.6  & 12.4 & 24.6 & 7.8 & 10.0 & 72.3 & 10.1 & \textbf{9.4} & 21.4 & 18.6 & 4.0 & 17.0$\pm$17.7 \\

\hline
\hline

\multirow{4}{*}{DTN~\cite{liu2019deep}} & APCER & 1.0 & 0.0 & 0.7 & 24.5 & 58.6 & 0.5 & 3.8 & 73.2 & 13.2 & 12.4 & 17.0 & 17.0 & 0.2 & 17.1$\pm$23.3 \\
\cline{3-16}  & BPCER & 18.6 & 11.9 & 29.3 & 12.8 & 13.4 & 8.5 & 23.0 & 11.5 & 9.6 & 16.0 & 21.5 & 22.6 & 16.8 & 16.6 $\pm$6.2 \\
\cline{3-16}  & ACER & \textbf{9.8} & 6.0 & 15.0 & 18.7 & 36.0 & 4.5 & 7.7 & 48.1 & 11.4 & 14.2 & \textbf{19.3} & 19.8 & 8.5 & 16.8 $\pm$11.1 \\
\cline{3-16}  & EER & \textbf{10.0} & \textbf{2.1} & 14.4 & 18.6 & 26.5 & \textbf{5.7} & 9.6 & 50.2 & 10.1 & 13.2 & \textbf{19.8} & 20.5 & 8.8 & 16.1$\pm$ 12.2 \\

\hline
\hline

\multirow{4}{*}{\textbf{Ours}} & APCER & 12.4 & 5.2 & 8.3 & 9.7 & 13.6 & 0.0 & 2.5 & 30.4 & 0.0 & 12.0 & 22.6 & 15.9 & 1.2 & 10.3$\pm$9.1 \\
\cline{3-16}  & BPCER & 13.2 & 6.2 & 13.1 & 10.8 & 16.3 & 3.9 & 2.3 & 34.1 & 1.6 & 13.9 & 23.2 & 17.1 & 2.3 & 12.2$\pm$9.4 \\
\cline{3-16}  & ACER & 12.8 & \textbf{5.7} & \textbf{10.7} & \textbf{10.3} & \textbf{14.9} & \textbf{1.9} & \textbf{2.4} & \textbf{32.3} & \textbf{0.8} & 12.9 & 22.9 & \textbf{16.5} & \textbf{1.7} & \textbf{11.2$\pm$9.2} \\
\cline{3-16}  & EER & 13.4 & 5.2 & \textbf{8.3}  & \textbf{9.7} & \textbf{13.6} & 5.8 & \textbf{2.5} & \textbf{33.8} & \textbf{0.0} & 14.0 & 23.3 & \textbf{16.6} & \textbf{1.2} & \textbf{11.3$\pm$9.5} \\

\hline
\hline

\end{tabular}
}
\vspace{-1.2em}
\label{tab:SiW-M}
\end{table}

\begin{table}[t]
\centering
\caption{The results of cross-dataset testing between CASIA-MFSD and Replay-Attack. The evaluation metric is HTER(\%). }
\vspace{-0.2em}
\scalebox{0.70}{\begin{tabular}{|c|c|c|c|c|}
\hline
\multirow{3}{*}{Method} &\multicolumn{2}{c|}{Protocol CR} &\multicolumn{2}{c|}{Protocol RC}\\
\cline{2-3}\cline{4-5}
&Train &Test &Train &Test\\
\cline{2-3} \cline{4-5} &\tabincell{c}{CASIA-MFSD} &\tabincell{c}{Replay-Attack} &\tabincell{c}{Replay-Attack} &\tabincell{c}{CASIA-MFSD}\\

\hline
LBP-TOP ~\cite{Pereira2013Can}
&\multicolumn{2}{c|}{49.7} &\multicolumn{2}{c|}{60.6} \\
\hline
STASN ~\cite{yang2019face}
&\multicolumn{2}{c|}{31.5} &\multicolumn{2}{c|}{30.9} \\
\hline
Color Texture ~\cite{Boulkenafet2017Face}
&\multicolumn{2}{c|}{30.3} &\multicolumn{2}{c|}{37.7} \\
\hline
FaceDs ~\cite{jourabloo2018face}
&\multicolumn{2}{c|}{28.5} &\multicolumn{2}{c|}{41.1} \\
\hline
Auxiliary ~\cite{Liu2018Learning}
&\multicolumn{2}{c|}{27.6} &\multicolumn{2}{c|}{28.4} \\
\hline
BASN ~\cite{kim2019basn}
&\multicolumn{2}{c|}{23.6} &\multicolumn{2}{c|}{29.9} \\
\hline
FAS-TD ~\cite{wang2018exploiting}
&\multicolumn{2}{c|}{17.5} &\multicolumn{2}{c|}{\textbf{24.0}} \\
\hline
\textbf{Ours}
&\multicolumn{2}{c|}{\textbf{16.6}} &\multicolumn{2}{c|}{36.4} \\
\hline
\end{tabular}
}
\label{tab:cross-testing}
\vspace{-0.6em}
\end{table}

\begin{table}[!htbp]
\centering
\vspace{-0.2em}
\caption{Fine-grained material recognition in SiW-M dataset. The evaluation metric is accuracy (\%).}
\vspace{-0.2em}
\scalebox{0.80}{\begin{tabular}{|c|c|c|c|c|c|c|}
\hline

Method & \ Live \  & \ Replay\  & \ Print\  & \ Mask\  & \ Makeup\  & \ Overall\  \\

\hline
ResNet50 (pre-trained)~\cite{he2016deep}
& 88.4 & 93.9 & 84.2 & 92.6 & \textbf{98.6} & 91.3  \\
\hline
Patch (Ours)
& 91.5 & 98.0 & 87.7 & 95.9 & 73.9 & 90.1 \\
\hline
Patch+BCN (Ours) 
& 83.7 & \textbf{100.0} & \textbf{93.0} & \textbf{96.0} & 94.2 & 92.0 \\
\hline
Patch+BCN+MFRM (Ours)
& \textbf{96.1} & \textbf{100.0} & \textbf{93.0} & 95.1 & 82.6 & \textbf{93.7} \\
\hline

\end{tabular}
}
\label{tab:material}
\vspace{-1.2em}
\end{table}

\textbf{Cross-dataset Testing.} \quad   In this experiment, we first train on the CASIA-MFSD and test on Replay-Attack, which is named as protocol CR. And then exchanging the training dataset and the testing dataset reciprocally, named protocol RC. As shown in Table~\ref{tab:cross-testing}, our proposed method has 16.6\% HTER on protocol CR, outperforming all prior state-of-the-arts. For protocol RC, we also achieve comparable performance with 36.4\% HTER. As our method is frame-level based, the performance might be further improved via introducing the temporal dynamic features in FAS-TD~\cite{wang2018exploiting}.

\begin{figure}[t]
\centering
\includegraphics[width=10.3cm,height=10cm]{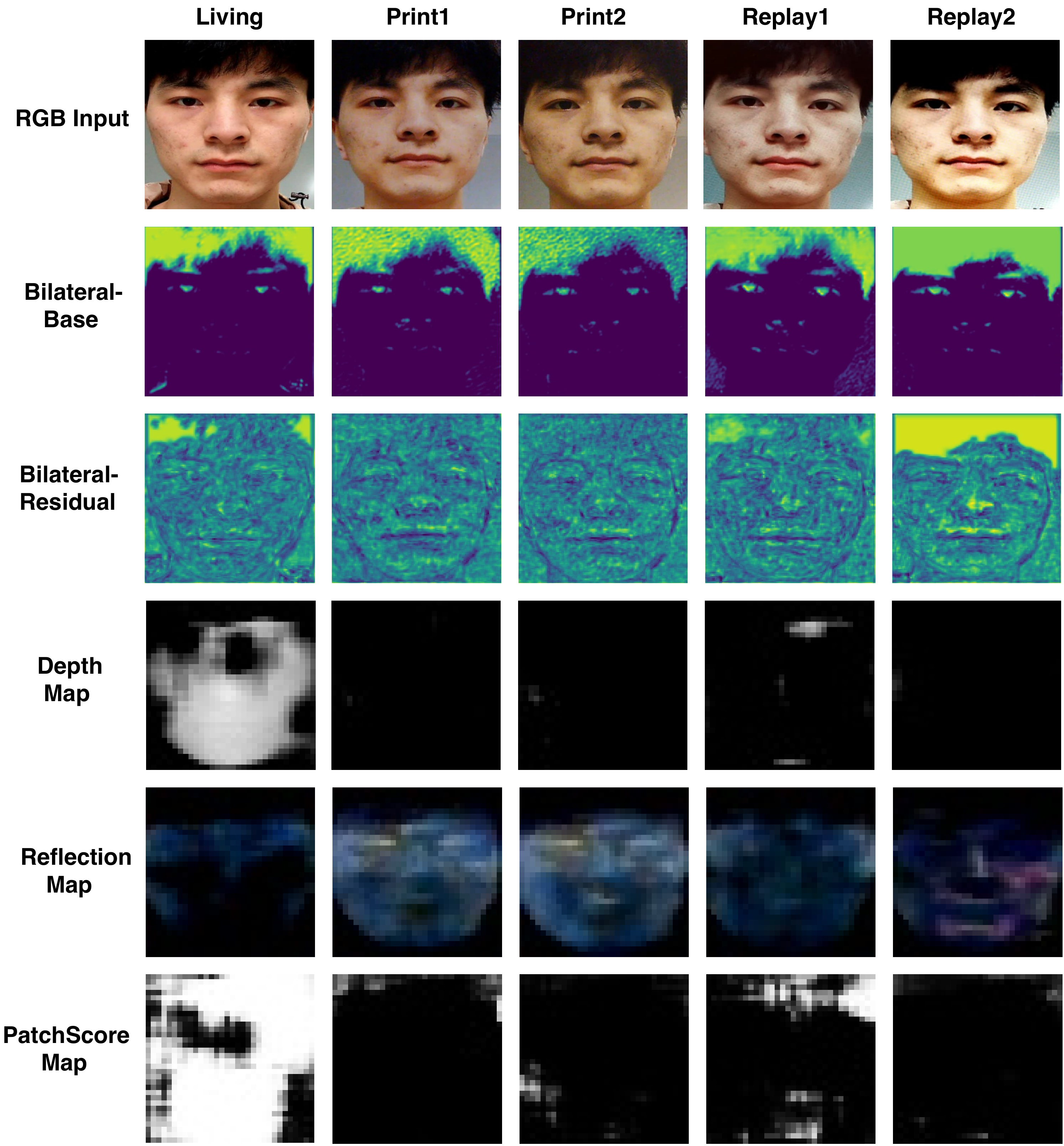}
  \caption{\small{
  Features visualization on live face (the first column) and spoofing faces (four columns to the right). The six rows represent the RGB images, low-level bilateral base features, low-level bilateral residual features in BCN, predicted depth maps, reflection maps and patch maps, respectively. Best view when zoom in.
  }
  }
\label{fig:visualization}
\vspace{-0.5em}
\end{figure}

 \vspace{-0.5em}
\subsection{Analysis and Visualization.}
 \vspace{-0.25em}
 \label{sec:Analysis}

In this subsection, we firstly conduct fine-grained material recognition on FAS dataset to prove that the learned features are material-based and intrinsic. And then we visualize and analyze the learned features.
 
\textbf{Fine-grained Material Recognition.}  Face anti-spoofing is usually regarded as a binary classification problem despite we treat it as a binary material perception problem (i.e., structural facial skin versus others) in this paper. It is curious whether the model learns material-based intrinsic features despite it achieves state-of-the-art performance in most FAS datasets. As there are rich spoofing material types in SiW-M dataset, we separate it into five material categories, i.e., live, replay, print, mask and makeup, which are made of structural facial skin, plain glass, wrapped paper, structural fiber and foundation, respectively. Half samples of the each category are used for training and the remaining parts are utilized for testing. Only patch map supervision with five categories is utilized as the baseline because depth and reflection maps are not suitable for multi-category classification. As shown in Table~\ref{tab:material}, with the BCN and MFRM, the overall accuracy boosts by 1.9\% and extra 1.7\% respectively, outperforming ResNet50 ~\cite{he2016deep} pre-trained in ImageNet. It implies intrinsic patterns among materials might be captured by BCN and MFRM.

\textbf{Features Visualization.} 
The low-level features of BCN and the predicted maps are visualized in Fig.~\ref{fig:visualization}. It is clear that the bilateral base and residual features between live and spoofing faces are quite different. For the bilateral base features of print attacks (see the 2nd and 3rd column in Fig.~\ref{fig:visualization}), the random noises in the hair region are obvious which is caused by the rough surface of the paper material. Moreover, the micro- patterns in bilateral residual features reveal more details of facial outline in spoofing attacks but blurriness in live faces.

\vspace{-1.5em}
\section{Conclusions}
\vspace{-0.5em}
\label{sec:conc}

In this paper, we rephrase face anti-spoofing (FAS) task as a material recognition problem and combine FAS with classical human material perception~\cite{sharan2013recognizing}. To this end, Bilateral Convolutional Networks are proposed for capturing material-based bilateral macro- and micro- features. Extensive experiments are performed to verify the effectiveness of the proposed method. Our future works include: 1) to learn intrinsic material features via disentangling them with material-unrelated features (e.g., face id and face attribute features); and 2) to establish a more suitable cross-material based FAS benchmark.

\textbf{Acknowledgment}  \quad This work was supported by the Academy of Finland for project MiGA (grant 316765), ICT 2023 project (grant 328115), and Infotech Oulu. As well, the authors wish to acknowledge CSC – IT Center for Science, Finland, for computational resources.


\section{Appendix}

\textbf{A. Impact of Spatial Neighborhood Distance in DBO} 

\begin{figure}
\vspace{-1.8em}
\centering
\includegraphics[width=8.0cm,height=2.7cm]{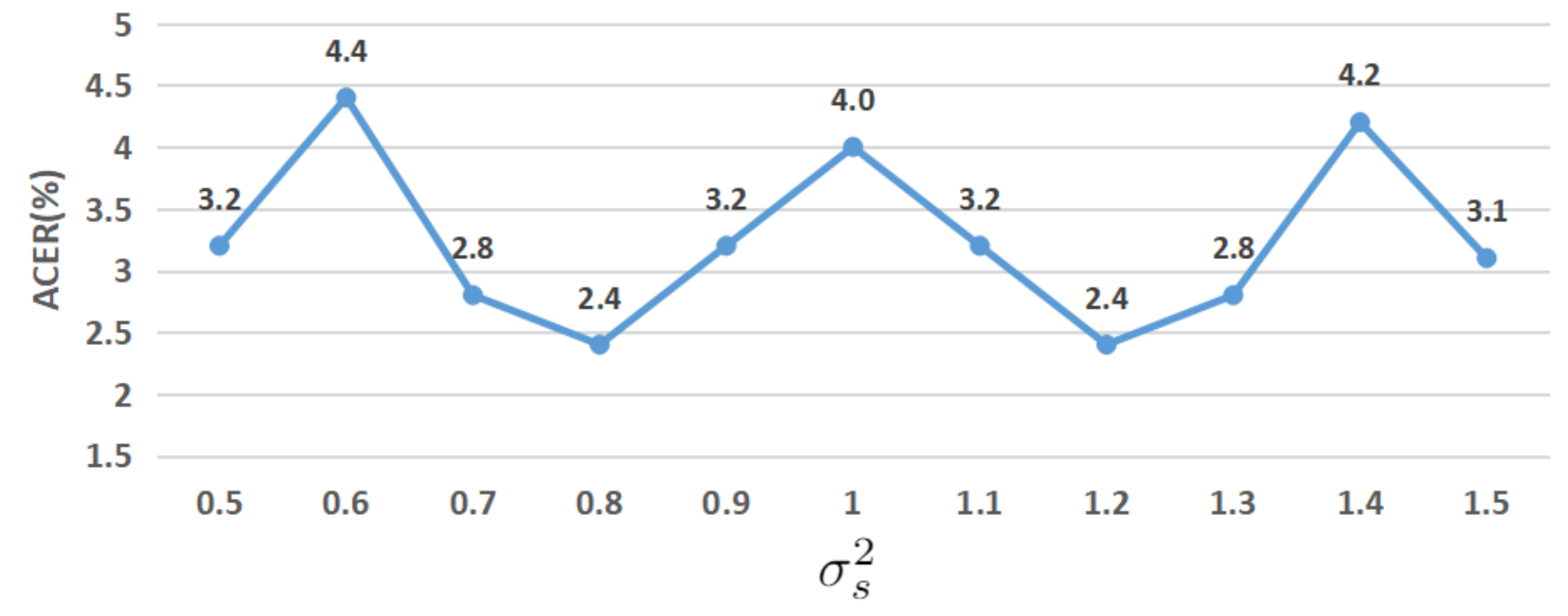}
  \caption{\small{
  Impact of $\sigma_{s}$ in BCN. 
  }
  }
\label{fig:AppendixA}
\vspace{-0.5em}
\end{figure}

The full version of deep bilateral operator (DBO) with spatial neighborhood distance term can be formulated as
\vspace{-0.5em}
\begin{equation} \small
\setlength{\belowdisplayskip}{-0.1em}
\begin{split}
& DBO_{full} (\mathcal{F})_{p}=\frac{1}{k}\sum_{q\in \mathcal{F}}g_{\sigma_{s}}(\left \| p-q \right \|)g_{\sigma_{r}}(\left | \mathcal{F}_{p}-\mathcal{F}_{q} \right |)\mathcal{F}_{q},
\\
&with: \qquad k=\sum_{q\in \mathcal{F}}g_{\sigma_{s}}(\left \| p-q \right \|)g_{\sigma_{r}}(\left | \mathcal{F}_{p}-\mathcal{F}_{q} \right |).
\end{split}
\label{eq:DBOfull}
\vspace{-0.2em}
\end{equation}
In this ablation study, the impact of spatial neighborhood distance $g_{\sigma_{s}}(\left \| p-q \right \|)$ would be evaluated. Here the default setting $\sigma_{r}^2=1.0$ is utilized. It can be seen from Fig.~\ref{fig:AppendixA} that there are no improvements (2.4\% and 2.1\% ACER for with and without spatial neighborhood distance, respectively) when introducing spatial neighborhood distance term into BCN.

Note that the ablation study about distance term $\sigma_{s}$ here is not enough thus it might be a sub-optimal solution. The optimal hyperparameter setting could be found via strict grid search, which is one of our future works. Long-range spatial impact of distance term $\sigma_{s}$ under large kernel size (e.g., 5x5 and 7x7) is also worth exploring in future.

\noindent\textbf{B. Network Details of Multi-head Supervision} 

\begin{figure}[htb]
\centering
\includegraphics[scale=0.37]{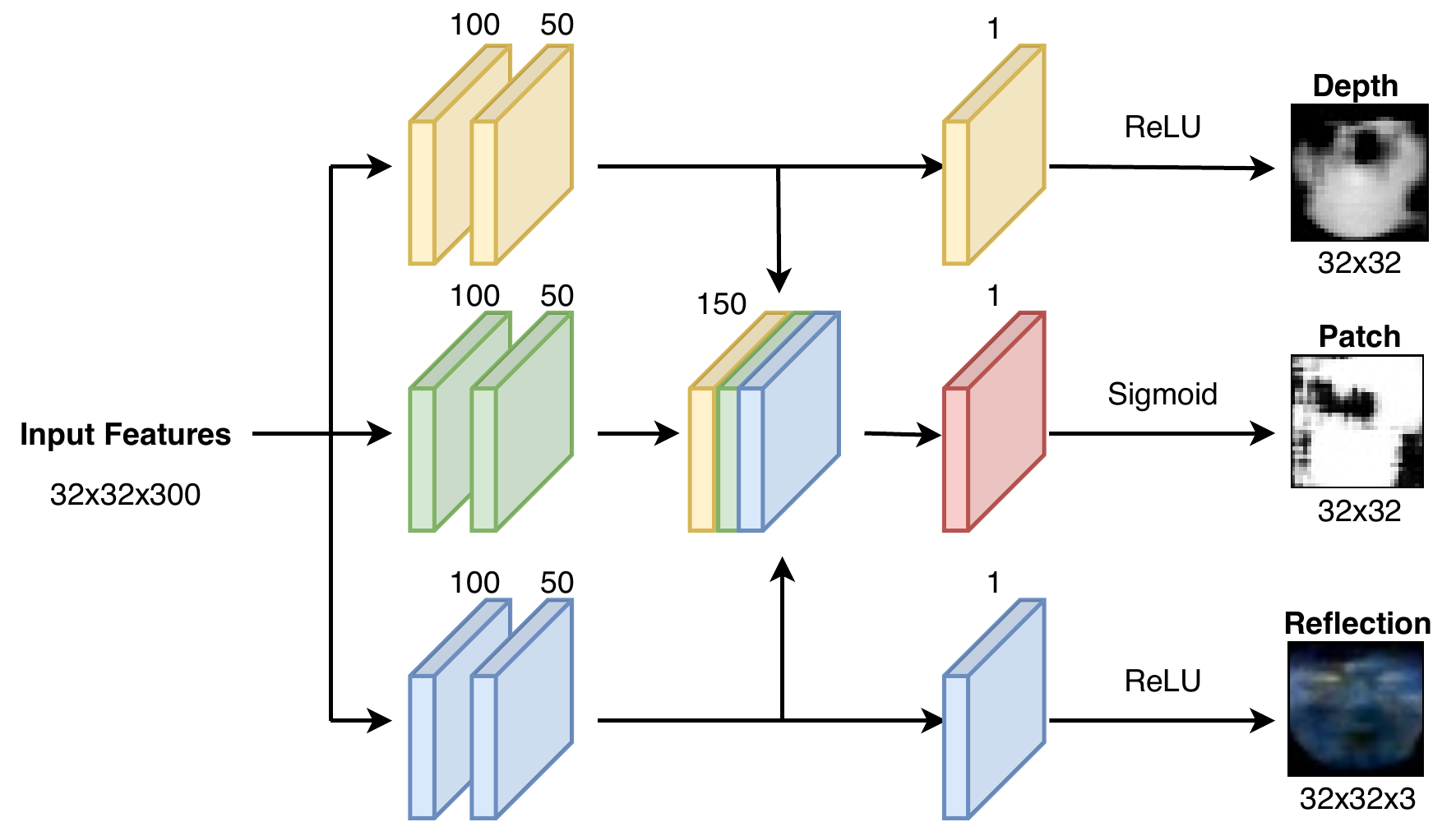}
  \caption{\small{
  Network structure of multi-head supervision. The number of filters are shown on top of each convolution, the size of all filters is 3$\times$3 with stride 1. 
  }
  }
\label{fig:multihead}
\vspace{-0.5em}
\end{figure}

The detailed convolutional layers are illustrated in Fig.~\ref{fig:multihead}. With the supervision from three kinds of cues, the backbone network is able to learn more holistic material-based features.

\vspace{-1.8em}

\noindent\textbf{C. Intra Testing Results on SiW} 

As shown in Table~\ref{tab:SiW}, the proposed method performs the best for all three protocols, revealing the excellent generalization capacity.

\begin{table}[htb]
\centering
\caption{The results of intra testing on three protocols of SiW~\cite{Liu2018Learning}. } 
\resizebox{0.56\textwidth}{!}{
\begin{tabular}{|c|c|c|c|c|}
\hline
Prot. & Method & APCER(\%) & BPCER(\%) & ACER(\%) \\
\hline
\multirow{3}{*}{1} 
       &Auxiliary ~\cite{Liu2018Learning}&3.58 &3.58 &3.58 \\
       &STASN ~\cite{yang2019face}&\textbf{--} &\textbf{--} &1.00 \\
       &FAS-TD ~\cite{wang2018exploiting} &0.96 &0.50 &0.73 \\
        &BASN ~\cite{kim2019basn}&\textbf{--} &\textbf{--} & 0.37 \\
        &\textbf{Ours}&0.55 &0.17 & \textbf{0.36} \\
\hline
\multirow{3}{*}{2} &Auxiliary ~\cite{Liu2018Learning}&0.57$\pm$0.69 &0.57$\pm$0.69 &0.57$\pm$0.69 \\
       &STASN ~\cite{yang2019face}&\textbf{--} &\textbf{--} &0.28$\pm$0.05 \\
       &FAS-TD ~\cite{wang2018exploiting}&0.08$\pm$0.14 &0.21$\pm$0.14 & 0.15$\pm$0.14 \\
       &BASN ~\cite{kim2019basn}&\textbf{--} &\textbf{--} &0.12$\pm$0.03 \\
       &\textbf{Ours} &0.08$\pm$0.17 &0.15$\pm$0.00  & \textbf{0.11$\pm$0.08} \\
\hline
\multirow{3}{*}{3} &STASN ~\cite{yang2019face}&\textbf{--} &\textbf{--} &12.10$\pm$1.50 \\
       &Auxiliary ~\cite{Liu2018Learning}&8.31$\pm$3.81 &8.31$\pm$3.80 &8.31$\pm$3.81 \\
       &BASN ~\cite{kim2019basn}&\textbf{--} &\textbf{--} &6.45$\pm$1.80 \\
       &FAS-TD ~\cite{wang2018exploiting}&3.10$\pm$0.81 &3.09$\pm$0.81 & 3.10$\pm$0.81 \\
       &\textbf{Ours} &2.55$\pm$0.89 &2.34$\pm$0.47  & \textbf{2.45$\pm$0.68} \\
\hline
\end{tabular}
}
\label{tab:SiW}
\vspace{-1.0em}
\end{table}

\noindent\textbf{D. Cross-type Testing on CASIA-MFSD, Replay-Attack and MSU-MFSD} 

In these cross-type testing, three datasets CASIA-MFSD~\cite{Zhang2012A}, Replay-Attack~\cite{ReplayAttack} and MSU-MFSD~\cite{wen2015face} are utilized to perform intra-dataset cross-type testing between replay and print attacks. For instance, the second column `Video' in Table~\ref{tab:cross-type} means that model should be trained from `Cut Photo' and `Wrapped Photo' while tested on `Video'. Table~\ref{tab:cross-type} shows that our proposed method achieves the best overall performance (96.77\% AUC), indicating the learned features generalized well among unknown attacks.

\begin{table}[htb]
\centering
\caption{Cross-type testing on CASIA-MFSD, Replay-Attack, and MSU-MFSD. The evaluation metric is AUC (\%).}

\scalebox{0.60}{\begin{tabular}{|c|c|c|c|c|c|c|c|c|c|c|}
\hline
\multirow{2}{*}{Method} &\multicolumn{3}{c|}{CASIA-MFSD ~\cite{Zhang2012A}} &\multicolumn{3}{c|}{Replay-Attack ~\cite{ReplayAttack}}&\multicolumn{3}{c|}{MSU-MFSD ~\cite{wen2015face}} &\multirow{2}{*}{Overall} \\
\cline{2-10} &\tabincell{c}{Video} &\tabincell{c}{Cut Photo} &\tabincell{c}{Wrapped Photo} &\tabincell{c}{Video}&\tabincell{c}{Digital Photo}&\tabincell{c}{Printed Photo}&\tabincell{c}{Printed Photo}&\tabincell{c}{HR Video}&\tabincell{c}{Mobile Video} & \\
\hline
OC-SVM+BSIF~\cite{arashloo2017anomaly}
& 70.74 & 60.73 & 95.90 & 84.03 & 88.14 & 73.66 & 64.81 & 87.44 & 74.69 & 78.68$\pm$11.74 \\
\hline
SVM+LBP ~\cite{Boulkenafet2017OULU}
& 91.94 & 91.70 & 84.47 & 99.08 & 98.17 & 87.28 & 47.68 & 99.50
 & 97.61 & 88.55$\pm$16.25 \\
\hline
NN+LBP ~\cite{xiong2018unknown}
& 94.16 & 88.39 & 79.85 & 99.75 & 95.17 & 78.86 & 50.57 & 99.93 & 93.54 & 86.69$\pm$16.25 \\
\hline
DTN ~\cite{liu2019deep}
& 90.0 & 97.3 & 97.5 & 99.9 & \textbf{99.9} & 99.6 & \textbf{81.6} & 99.9 & 97.5 & 95.9$\pm$6.2 \\
\hline
AIM-FAS ~\cite{qin2019learning}
& 93.6 & 99.7 & 99.1 & 99.8 & \textbf{99.9} & 99.8 & 76.3 & 99.9 & 99.1 & 96.4$\pm$7.8 \\
\hline
\textbf{Ours}
& \textbf{99.62} & \textbf{100.00} & \textbf{100.00} & \textbf{99.99} & 99.74 & \textbf{99.91} & 71.64 & \textbf{100.00} & \textbf{99.99} & \textbf{96.77$\pm$9.99} \\
\hline
\end{tabular}
}
\label{tab:cross-type}
\vspace{-0.2em}
\end{table}


\clearpage

\bibliographystyle{splncs}
\bibliography{egbib}

\begin{thebibliography}{10}

\bibitem{sharan2013recognizing}
Sharan, L., Liu, C., Rosenholtz, R., Adelson, E.H.:
\newblock Recognizing materials using perceptually inspired features.
\newblock International journal of computer vision \textbf{103}(3) (2013)
  348--371

\bibitem{boulkenafet2015face}
Boulkenafet, Z., Komulainen, J., Hadid, A.:
\newblock Face anti-spoofing based on color texture analysis.
\newblock In: IEEE international conference on image processing (ICIP). (2015)
  2636--2640

\bibitem{Boulkenafet2017Face}
Boulkenafet, Z., Komulainen, J., Hadid, A.:
\newblock Face spoofing detection using colour texture analysis.
\newblock IEEE Transactions on Information Forensics and Security
  \textbf{11}(8) (2016)  1818--1830

\bibitem{Pereira2012LBP}
de~Freitas~Pereira, T., Anjos, A., De~Martino, J.M., Marcel, S.:
\newblock Lbp- top based countermeasure against face spoofing attacks.
\newblock In: Asian Conference on Computer Vision. (2012)  121--132

\bibitem{Komulainen2014Context}
Komulainen, J., Hadid, A., Pietikainen, M.:
\newblock Context based face anti-spoofing.
\newblock In: 2013 IEEE Sixth International Conference on Biometrics: Theory,
  Applications and Systems (BTAS). (2013)  1--8

\bibitem{Peixoto2011Face}
Peixoto, B., Michelassi, C., Rocha, A.:
\newblock Face liveness detection under bad illumination conditions.
\newblock In: ICIP, IEEE (2011)  3557--3560

\bibitem{Patel2016Secure}
Patel, K., Han, H., Jain, A.K.:
\newblock Secure face unlock: Spoof detection on smartphones.
\newblock IEEE transactions on information forensics and security
  \textbf{11}(10) (2016)  2268--2283

\bibitem{qin2019learning}
Qin, Y., Zhao, C., Zhu, X., Wang, Z., Yu, Z., Fu, T., Zhou, F., Shi, J., Lei,
  Z.:
\newblock Learning meta model for zero-and few-shot face anti-spoofing.
\newblock The Thirty-Fourth AAAI Conference on Artificial Intelligence (AAAI)
  (2020)

\bibitem{wang2020deep}
Wang, Z., Yu, Z., Zhao, C., Zhu, X., Qin, Y., Zhou, Q., Zhou, F., Lei, Z.:
\newblock Deep spatial gradient and temporal depth learning for face
  anti-spoofing.
\newblock In: Proceedings of the IEEE/CVF Conference on Computer Vision and
  Pattern Recognition. (2020)  5042--5051

\bibitem{yu2020searching}
Yu, Z., Zhao, C., Wang, Z., Qin, Y., Su, Z., Li, X., Zhou, F., Zhao, G.:
\newblock Searching central difference convolutional networks for face
  anti-spoofing.
\newblock In: Proceedings of the IEEE/CVF Conference on Computer Vision and
  Pattern Recognition. (2020)  5295--5305

\bibitem{jourabloo2018face}
Jourabloo, A., Liu, Y., Liu, X.:
\newblock Face de-spoofing: Anti-spoofing via noise modeling.
\newblock In: Proceedings of the European Conference on Computer Vision (ECCV).
  (2018)  290--306

\bibitem{lin2019face}
Lin, B., Li, X., Yu, Z., Zhao, G.:
\newblock Face liveness detection by rppg features and contextual patch-based
  cnn.
\newblock In: Proceedings of the 2019 3rd International Conference on Biometric
  Engineering and Applications, ACM (2019)  61--68

\bibitem{li2016generalized}
Li, X., Komulainen, J., Zhao, G., Yuen, P.C., Pietik{\"a}inen, M.:
\newblock Generalized face anti-spoofing by detecting pulse from face videos.
\newblock In: 2016 23rd International Conference on Pattern Recognition (ICPR),
  IEEE (2016)  4244--4249

\bibitem{Liu2018Learning}
Liu, Y., Jourabloo, A., Liu, X.:
\newblock Learning deep models for face anti-spoofing: Binary or auxiliary
  supervision.
\newblock In: Proceedings of the IEEE Conference on Computer Vision and Pattern
  Recognition. (2018)  389--398

\bibitem{liu2018remote}
Liu, S.Q., Lan, X., Yuen, P.C.:
\newblock Remote photoplethysmography correspondence feature for 3d mask face
  presentation attack detection.
\newblock In: Proceedings of the European Conference on Computer Vision (ECCV).
  (2018)  558--573

\bibitem{Atoum2018Face}
Atoum, Y., Liu, Y., Jourabloo, A., Liu, X.:
\newblock Face anti-spoofing using patch and depth-based cnns.
\newblock In: 2017 IEEE International Joint Conference on Biometrics (IJCB).
  (2017)  319--328

\bibitem{tan2010face}
Tan, X., Li, Y., Liu, J., Jiang, L.:
\newblock Face liveness detection from a single image with sparse low rank
  bilinear discriminative model.
\newblock In: European Conference on Computer Vision, Springer (2010)  504--517

\bibitem{li20193d1}
Li, L., Xia, Z., Jiang, X., Ma, Y., Roli, F., Feng, X.:
\newblock 3d face mask presentation attack detection based on intrinsic image
  analysis.
\newblock arXiv preprint arXiv:1903.11303 (2019)

\bibitem{Boulkenafet2017Face_SURF}
Boulkenafet, Z., Komulainen, J., Hadid, A.:
\newblock Face antispoofing using speeded-up robust features and fisher vector
  encoding.
\newblock IEEE Signal Processing Letters \textbf{24}(2) (2017)  141--145

\bibitem{komulainen2012face}
Komulainen, J., Hadid, A., Pietik{\"a}inen, M.:
\newblock Face spoofing detection using dynamic texture.
\newblock In: Asian Conference on Computer Vision, Springer (2012)  146--157

\bibitem{siddiqui2016face}
Siddiqui, T.A., Bharadwaj, S., Dhamecha, T.I., Agarwal, A., Vatsa, M., Singh,
  R., Ratha, N.:
\newblock Face anti-spoofing with multifeature videolet aggregation.
\newblock In: 2016 23rd International Conference on Pattern Recognition (ICPR),
  IEEE (2016)  1035--1040

\bibitem{Pan2007Eyeblink}
Pan, G., Sun, L., Wu, Z., Lao, S.:
\newblock Eyeblink-based anti-spoofing in face recognition from a generic
  webcamera.
\newblock In: IEEE International Conference on Computer Vision. (2007)  1--8

\bibitem{yu2020auto}
Yu, Z., Qin, Y., Xu, X., Zhao, C., Wang, Z., Lei, Z., Zhao, G.:
\newblock Auto-fas: Searching lightweight networks for face anti-spoofing.
\newblock In: ICASSP 2020-2020 IEEE International Conference on Acoustics,
  Speech and Signal Processing (ICASSP), IEEE (2020)  996--1000

\bibitem{Li2017An}
Li, L., Feng, X., Boulkenafet, Z., Xia, Z., Li, M., Hadid, A.:
\newblock An original face anti-spoofing approach using partial convolutional
  neural network.
\newblock In: IPTA. (2016)  1--6

\bibitem{Patel2016Cross}
Patel, K., Han, H., Jain, A.K.:
\newblock Cross-database face antispoofing with robust feature representation.
\newblock In: Chinese Conference on Biometric Recognition. (2016)  611--619

\bibitem{george2019deep}
George, A., Marcel, S.:
\newblock Deep pixel-wise binary supervision for face presentation attack
  detection.
\newblock In: International Conference on Biometrics. Number CONF (2019)

\bibitem{yu2020multi}
Yu, Z., Qin, Y., Li, X., Wang, Z., Zhao, C., Lei, Z., Zhao, G.:
\newblock Multi-modal face anti-spoofing based on central difference networks.
\newblock In: Proceedings of the IEEE/CVF Conference on Computer Vision and
  Pattern Recognition Workshops. (2020)  650--651

\bibitem{kim2019basn}
Kim, T., Kim, Y., Kim, I., Kim, D.:
\newblock Basn: Enriching feature representation using bipartite auxiliary
  supervisions for face anti-spoofing.
\newblock In: Proceedings of the IEEE International Conference on Computer
  Vision Workshops. (2019)  0--0

\bibitem{liu2019deep}
Liu, Y., Stehouwer, J., Jourabloo, A., Liu, X.:
\newblock Deep tree learning for zero-shot face anti-spoofing.
\newblock In: Proceedings of the IEEE Conference on Computer Vision and Pattern
  Recognition. (2019)  4680--4689

\bibitem{jia2020single}
Jia, Y., Zhang, J., Shan, S., Chen, X.:
\newblock Single-side domain generalization for face anti-spoofing.
\newblock In: Proceedings of the IEEE/CVF Conference on Computer Vision and
  Pattern Recognition. (2020)  8484--8493

\bibitem{wang2020cross}
Wang, G., Han, H., Shan, S., Chen, X.:
\newblock Cross-domain face presentation attack detection via multi-domain
  disentangled representation learning.
\newblock In: Proceedings of the IEEE/CVF Conference on Computer Vision and
  Pattern Recognition. (2020)  6678--6687

\bibitem{shao2019multi}
Shao, R., Lan, X., Li, J., Yuen, P.C.:
\newblock Multi-adversarial discriminative deep domain generalization for face
  presentation attack detection.
\newblock In: Proceedings of the IEEE Conference on Computer Vision and Pattern
  Recognition. (2019)  10023--10031

\bibitem{wang2018exploiting}
Wang, Z., Zhao, C., Qin, Y., Zhou, Q., Lei, Z.:
\newblock Exploiting temporal and depth information for multi-frame face
  anti-spoofing.
\newblock arXiv preprint arXiv:1811.05118 (2018)

\bibitem{yang2019face}
Yang, X., Luo, W., Bao, L., Gao, Y., Gong, D., Zheng, S., Li, Z., Liu, W.:
\newblock Face anti-spoofing: Model matters, so does data.
\newblock In: Proceedings of the IEEE Conference on Computer Vision and Pattern
  Recognition. (2019)

\bibitem{lin2018live}
Lin, C., Liao, Z., Zhou, P., Hu, J., Ni, B.:
\newblock Live face verification with multiple instantialized local homographic
  parameterization.
\newblock In: IJCAI. (2018)  814--820

\bibitem{yu2019remote}
Yu, Z., Li, X., Zhao, G.:
\newblock Remote photoplethysmograph signal measurement from facial videos
  using spatio-temporal networks.
\newblock arXiv preprint arXiv:1905.02419 (2019)

\bibitem{yu2020autohr}
Yu, Z., Li, X., Niu, X., Shi, J., Zhao, G.:
\newblock Autohr: A strong end-to-end baseline for remote heart rate
  measurement with neural searching.
\newblock arXiv preprint arXiv:2004.12292 (2020)

\bibitem{yu2019remote2}
Yu, Z., Peng, W., Li, X., Hong, X., Zhao, G.:
\newblock Remote heart rate measurement from highly compressed facial videos:
  an end-to-end deep learning solution with video enhancement.
\newblock In: Proceedings of the IEEE International Conference on Computer
  Vision. (2019)  151--160

\bibitem{maloney2010color}
Maloney, L.T., Brainard, D.H.:
\newblock Color and material perception: Achievements and challenges.
\newblock Journal of Vision \textbf{10}(9) (2010)  19--19

\bibitem{fleming2014visual}
Fleming, R.W.:
\newblock Visual perception of materials and their properties.
\newblock Vision research \textbf{94} (2014)  62--75

\bibitem{nishida2019image}
Nishida, S.:
\newblock Image statistics for material perception.
\newblock Current Opinion in Behavioral Sciences \textbf{30} (2019)  94--99

\bibitem{adelson2001seeing}
Adelson, E.H.:
\newblock On seeing stuff: the perception of materials by humans and machines.
\newblock In: Human vision and electronic imaging VI. Volume 4299.,
  International Society for Optics and Photonics (2001)  1--12

\bibitem{varma2008statistical}
Varma, M., Zisserman, A.:
\newblock A statistical approach to material classification using image patch
  exemplars.
\newblock IEEE transactions on pattern analysis and machine intelligence
  \textbf{31}(11) (2008)  2032--2047

\bibitem{jiang2018deep}
Jiang, X., Du, J., Sun, B., Feng, X.:
\newblock Deep dilated convolutional network for material recognition.
\newblock In: 2018 Eighth International Conference on Image Processing Theory,
  Tools and Applications (IPTA), IEEE (2018)  1--6

\bibitem{ling2018role}
Ling, S., Callet, P.L., Yu, Z.:
\newblock The role of structure and textural information in image utility and
  quality assessment tasks.
\newblock Electronic Imaging \textbf{2018}(14) (2018)  1--13

\bibitem{deng2016video}
Deng, B.W., Yu, Z.T., Ling, B.W., Yang, Z.:
\newblock Video quality assessment based on features for semantic task and
  human material perception.
\newblock In: 2016 IEEE International Conference on Consumer Electronics-China
  (ICCE-China), IEEE (2016)  1--4

\bibitem{li20203d}
Li, L., Xia, Z., Jiang, X., Ma, Y., Roli, F., Feng, X.:
\newblock 3d face mask presentation attack detection based on intrinsic image
  analysis.
\newblock IET Biometrics \textbf{9}(3) (2020)  100--108

\bibitem{tomasi1998bilateral}
Tomasi, C., Manduchi, R.:
\newblock Bilateral filtering for gray and color images.
\newblock In: Iccv. Volume~98. (1998) ~2

\bibitem{durand2002fast}
Durand, F., Dorsey, J.:
\newblock Fast bilateral filtering for the display of high-dynamic-range
  images.
\newblock In: ACM transactions on graphics (TOG). Volume~21., ACM (2002)
  257--266

\bibitem{paris2006fast}
Paris, S., Durand, F.:
\newblock A fast approximation of the bilateral filter using a signal
  processing approach.
\newblock In: European conference on computer vision, Springer (2006)  568--580

\bibitem{he2016deep}
He, K., Zhang, X., Ren, S., Sun, J.:
\newblock Deep residual learning for image recognition.
\newblock In: Proceedings of the IEEE conference on computer vision and pattern
  recognition. (2016)  770--778

\bibitem{wang2019carafe}
Wang, J., Chen, K., Xu, R., Liu, Z., Loy, C.C., Lin, D.:
\newblock Carafe: Content-aware reassembly of features.
\newblock arXiv preprint arXiv:1905.02188 (2019)

\bibitem{woo2018cbam}
Woo, S., Park, J., Lee, J.Y., So~Kweon, I.:
\newblock Cbam: Convolutional block attention module.
\newblock In: Proceedings of the European Conference on Computer Vision (ECCV).
  (2018)  3--19

\bibitem{hu2018squeeze}
Hu, J., Shen, L., Sun, G.:
\newblock Squeeze-and-excitation networks.
\newblock In: Proceedings of the IEEE conference on computer vision and pattern
  recognition. (2018)  7132--7141

\bibitem{wang2018non}
Wang, X., Girshick, R., Gupta, A., He, K.:
\newblock Non-local neural networks.
\newblock In: Proceedings of the IEEE Conference on Computer Vision and Pattern
  Recognition. (2018)  7794--7803

\bibitem{Feng2018Joint}
Feng, Y., Wu, F., Shao, X., Wang, Y., Zhou, X.:
\newblock Joint 3d face reconstruction and dense alignment with position map
  regression network.
\newblock In: Proceedings of the European Conference on Computer Vision (ECCV).
  (2017)

\bibitem{zhang2018single}
Zhang, X., Ng, R., Chen, Q.:
\newblock Single image reflection separation with perceptual losses.
\newblock In: Proceedings of the IEEE Conference on Computer Vision and Pattern
  Recognition. (2018)  4786--4794

\bibitem{Boulkenafet2017OULU}
Boulkenafet, Z., Komulainen, J., Li, L., Feng, X., Hadid, A.:
\newblock Oulu-npu: A mobile face presentation attack database with real-world
  variations.
\newblock In: FGR. (2017)  612--618

\bibitem{Zhang2012A}
Zhang, Z., Yan, J., Liu, S., Lei, Z., Yi, D., Li, S.Z.:
\newblock A face antispoofing database with diverse attacks.
\newblock In: ICB. (2012)  26--31

\bibitem{ReplayAttack}
Chingovska, I., Anjos, A., Marcel, S.:
\newblock On the effectiveness of local binary patterns in face anti-spoofing.
\newblock In: Biometrics Special Interest Group. (2012)  1--7

\bibitem{wen2015face}
Wen, D., Han, H., Jain, A.K.:
\newblock Face spoof detection with image distortion analysis.
\newblock IEEE Transactions on Information Forensics and Security
  \textbf{10}(4) (2015)  746--761

\bibitem{ACER}
international organization~for standardization:
\newblock Iso/iec jtc 1/sc 37 biometrics: Information technology biometric
  presentation attack detection part 1: Framework.
\newblock In: https://www.iso.org/obp/ui/iso. (2016)

\bibitem{boulkenafet2017competition}
Boulkenafet, Z., Komulainen, J., Akhtar, Z., Benlamoudi, A., Samai, D.,
  Bekhouche, S.E., Ouafi, A., Dornaika, F., Taleb-Ahmed, A., Qin, L.,  et~al.:
\newblock A competition on generalized software-based face presentation attack
  detection in mobile scenarios.
\newblock In: 2017 IEEE International Joint Conference on Biometrics (IJCB),
  IEEE (2017)  688--696

\bibitem{arashloo2017anomaly}
Arashloo, S.R., Kittler, J., Christmas, W.:
\newblock An anomaly detection approach to face spoofing detection: A new
  formulation and evaluation protocol.
\newblock IEEE Access \textbf{5} (2017)  13868--13882

\bibitem{xiong2018unknown}
Xiong, F., AbdAlmageed, W.:
\newblock Unknown presentation attack detection with face rgb images.
\newblock In: 2018 IEEE 9th International Conference on Biometrics Theory,
  Applications and Systems (BTAS), IEEE (2018)  1--9

\bibitem{Pereira2013Can}
de~Freitas~Pereira, T., Anjos, A., De~Martino, J.M., Marcel, S.:
\newblock Can face anti-spoofing countermeasures work in a real world scenario?
\newblock In: 2013 international conference on biometrics (ICB). (2013)  1--8

\end{thebibliography}
\end{document}